\PassOptionsToPackage{table}{xcolor}
\documentclass[]{fairmeta}

\usepackage{amsmath,amsfonts,bm}

\usepackage{hyperref}
\usepackage{url}
\usepackage{booktabs}
\usepackage{tabularx}
\usepackage[table]{xcolor}
\definecolor{vistaClosed}{HTML}{F7EDE1}
\definecolor{vistaOpen}{HTML}{F0E9F8}
\definecolor{vistaAgent}{HTML}{DCEEE3}
\definecolor{vistaControl}{HTML}{F0F3F6}
\definecolor{vistaOurs}{HTML}{D6E5F4}
\definecolor{vistaHighlight}{HTML}{EDF4FB}
\definecolor{vistaCitation}{HTML}{4F7294}
\usepackage{multirow}
\usepackage{graphicx}
\usepackage{placeins}
\usepackage{listings}
\usepackage{amsmath}
\usepackage{amssymb}
\usepackage{microtype}
\usepackage{xspace}
\hypersetup{colorlinks=true,citecolor=vistaCitation,linkcolor=black,urlcolor=black,filecolor=black}

\hypersetup{pdftitle={VISTA: Internalizing Collective Visual Experience via On-Policy Distillation for Active Multimodal Agents},pdfsubject={Active multimodal agents and on-policy distillation}}

\title{VISTA: Internalizing Collective Visual Experience via On-Policy Distillation for Active Multimodal Agents}

\author{Zheng Jiang}
\author{Houde Qian}
\author{Yiming Chen}
\author{Ling Li}
\author{Chaoyang Li}
\author{Yueqi Li}
\author{Yuxuan Liu}
\author[\dagger]{Lifeng Sun}

\affiliation{Tsinghua University}
\contribution[\dagger]{Corresponding author}
\metadata[Code]{\url{https://github.com/jiangz20/VISTA}}
\metadata[Contact]{\email{jz24@mails.tsinghua.edu.cn}, \email{sunlf@tsinghua.edu.cn}}
\hypersetup{pdfauthor={Zheng Jiang, Houde Qian, Yiming Chen, Ling Li, Chaoyang Li, Yueqi Li, Yuxuan Liu, Lifeng Sun}}

\newcommand{\method}{\textsc{VISTA}\xspace}
\newcommand{\cved}{\textsc{CVED}\xspace}
\newcommand{\hapi}{\textsc{HAPI}\xspace}
\newcommand{\experience}{\mathcal{E}}

\newcommand{\traj}{\tau}
\newcommand{\student}{\pi_{\theta}}
\newcommand{\teacher}{\pi_{\mathrm{T}}}

\abstract{
Active multimodal agents use visual tools to acquire task-relevant evidence while reasoning.
Although reinforcement learning samples multiple interaction trajectories per input, outcome-based objectives primarily use the group to estimate scalar advantages, leaving complementary visual discoveries underused.
We introduce \method, which internalizes collective visual experience through on-policy distillation by turning observations from same-input rollouts into shared supervision.
\emph{Collective visual experience distillation} (\cved) organizes these observations with their interaction context and aligns them with individual decisions, while \emph{heterogeneity-aware policy improvement} (\hapi) reinforces successful trajectories and provides experience-guided distillation for unsuccessful attempts.
An experience-conditioned teacher evaluates the student's sampled response prefixes, allowing discoveries from one trajectory to guide learning in another without replacing the student's original history or generating new target trajectories.
The trained agent retains its visual tools and acts using its own interaction history.
\method achieves the strongest average performance among the evaluated active multimodal agents of comparable size and consistently outperforms same-backbone training baselines across fine-grained perception and general reasoning tasks, demonstrating the value of collective experience for active multimodal learning.
}

\begin{document}

\maketitle
\pagestyle{plain}
\thispagestyle{empty}

\section{Introduction}
\label{sec:introduction}

Multimodal agents are moving beyond answering questions from a fixed image toward actively acquiring the evidence required to solve them.
They can inspect local regions with crop and zoom tools, transform returned views, and chain visual actions with reasoning or search
\citep{wu2023vstar,hu2024visualsketchpad,shen2024zoomeye}.
This setting makes visual interaction part of the policy.
Success depends not only on what the model knows, but also on whether it acquires useful evidence and incorporates it into subsequent decisions.
Reinforcement learning has therefore become a natural training paradigm for active multimodal agents.
Recent systems sample multiple tool-using trajectories for each image--question pair and use verifiable outcomes to reinforce effective behavior
\citep{zheng2025deepeyes,wu2025vtoolr1,su2025pixelreasoner,liu2025visualarft,yang2026mapo,shao2024deepseekmath}.

However, current outcome-based objectives use a rollout group primarily as a source of scalar comparisons.
Once relative advantages have been computed, the visual discoveries made by one trajectory are not directly available to the others: a trajectory that finds a decisive region can teach its peers only through a reward difference.
This creates an \emph{experience-utilization gap}.
The group may collectively contain the evidence needed to solve a task, including useful observations from partial attempts, yet most of this task-specific experience disappears after the policy update.
On-policy distillation offers a natural way to narrow this gap by evaluating a student's response under an enhanced context and thereby providing denser token-level supervision
\citep{penaloza2026privileged,ye2026opcd,yang2026opid}.
Visual approaches derive supervision from privileged regional views, generated visual thoughts, or contrasts between visual inputs
\citep{yuan2026visionopd,cai2026imagineopd,li2026visualopsd,liang2026vcsd}.

Applying visual OPD to active multimodal agents nevertheless leaves two problems unresolved.
The first concerns the source of privileged supervision.
Existing approaches enrich teacher predictions through privileged context or visual contrast, but leave open how to exploit complementary evidence discovered across the agent's own same-input interaction rollouts.
Turning these discoveries into supervision requires organizing their interaction context and aligning them with the student's current decision.
The second problem concerns the allocation of supervision across heterogeneous trajectories.
A failed attempt may contain a useful observation despite reaching an incorrect answer, whereas a successful attempt may benefit more from reinforcing its behavior than from matching a teacher's preferences.
Final success therefore does not fully characterize the value of a trajectory's experience, and uniform distillation does not distinguish behavioral reinforcement from corrective guidance.
Effective learning therefore requires both a shared representation of the group's experience and a policy-improvement strategy that respects the heterogeneous roles of its trajectories.

These observations motivate VISTA.
Our first principle is to distill the shared evidence produced by the rollout group rather than treating trajectories as independent reward samples.
We implement this principle with \emph{collective visual experience distillation} (\cved), which organizes the group's visual artifacts and interaction context into training-only privilege and aligns that context with each student's original history.
Our second principle is to improve the policy according to the heterogeneous roles of different trajectories.
We implement it with \emph{heterogeneity-aware policy improvement} (\hapi): successful trajectories receive reward-aligned RLVR updates, while unsuccessful trajectories receive OPSD guidance from the experience-conditioned teacher.
The teacher evaluates the exact response prefix sampled by the student rather than generating a replacement trajectory.
The student remains conditioned only on its own interaction history, and the rollout group, teacher, and collective context are removed at deployment.
VISTA therefore transfers useful discoveries across trajectories during training while preserving the original active-agent interface at inference time.

Our contributions are:
\begin{itemize}
    \item \textbf{Mechanism.} We identify an experience-utilization gap in active multimodal learning, where outcome-based supervision underuses complementary visual discoveries and overlooks the distinction between trajectory success and informational value.
    \item \textbf{Methodology.} We introduce \method, combining \cved for decision-aligned experience distillation with \hapi for reinforcing successful trajectories and guiding unsuccessful ones through collective visual experience.
    \item \textbf{Performance.} \method leads the evaluated active multimodal agents of comparable size in average visual-reasoning accuracy and improves over same-backbone training baselines across fine-grained perception and general reasoning tasks.
\end{itemize}

\section{Related Work}
\label{sec:related_work}

\paragraph{RLVR for multimodal agents.}
Reinforcement learning with verifiable rewards enables multimodal agents to acquire evidence through visual tools rather than reason only over a fixed image \citep{zheng2025deepeyes,su2025pixelreasoner,liu2025visualarft}.
VTool-R1 and Thyme extend this paradigm to executable image operations, while DeepEyesV2 develops broader multimodal tool use \citep{wu2025vtoolr1,zhang2025thyme,hong2025deepeyesv2}.
Further work improves reasoning--action alignment and scales interaction depth \citep{yang2026mapo,lai2025minio3}.
These studies primarily improve how agents explore and learn from task feedback.
\method addresses the complementary problem of reusing what exploration discovers: visual observations from same-input rollouts become shared supervision for the agent's decisions.

\paragraph{OPD for multimodal agents.}
On-policy distillation trains on student-generated responses, and self-distillation can obtain stronger supervision by conditioning the same model on additional information \citep{agarwal2023gkd,zhao2026opsd}.
Vision-OPD and Imagine-OPD exploit regional evidence, while Visual-OPSD transfers the benefit of generated visual thoughts \citep{yuan2026visionopd,cai2026imagineopd,li2026visualopsd}.
VA-OPD, V-Zero, and VCSD use visual dependence or contrastive evidence to refine the distillation signal \citep{liu2026vaopd,sun2026vzero,liang2026vcsd}.
Context- and skill-conditioned approaches also distill trajectory experience for language models and agents \citep{ye2026opcd,yang2026opid}.
\method focuses on visual experience acquired collectively through active interaction: CVED aligns this evidence with student decisions, and HAPI distinguishes reinforcement of successful behavior from distillation on unsuccessful attempts.
Appendix~\ref{sec:additional_related_work} provides a detailed discussion.

\section{VISTA}
\label{sec:method}

\begin{figure}[!t]
\vspace{-1mm}
\centering
\includegraphics[width=\linewidth]{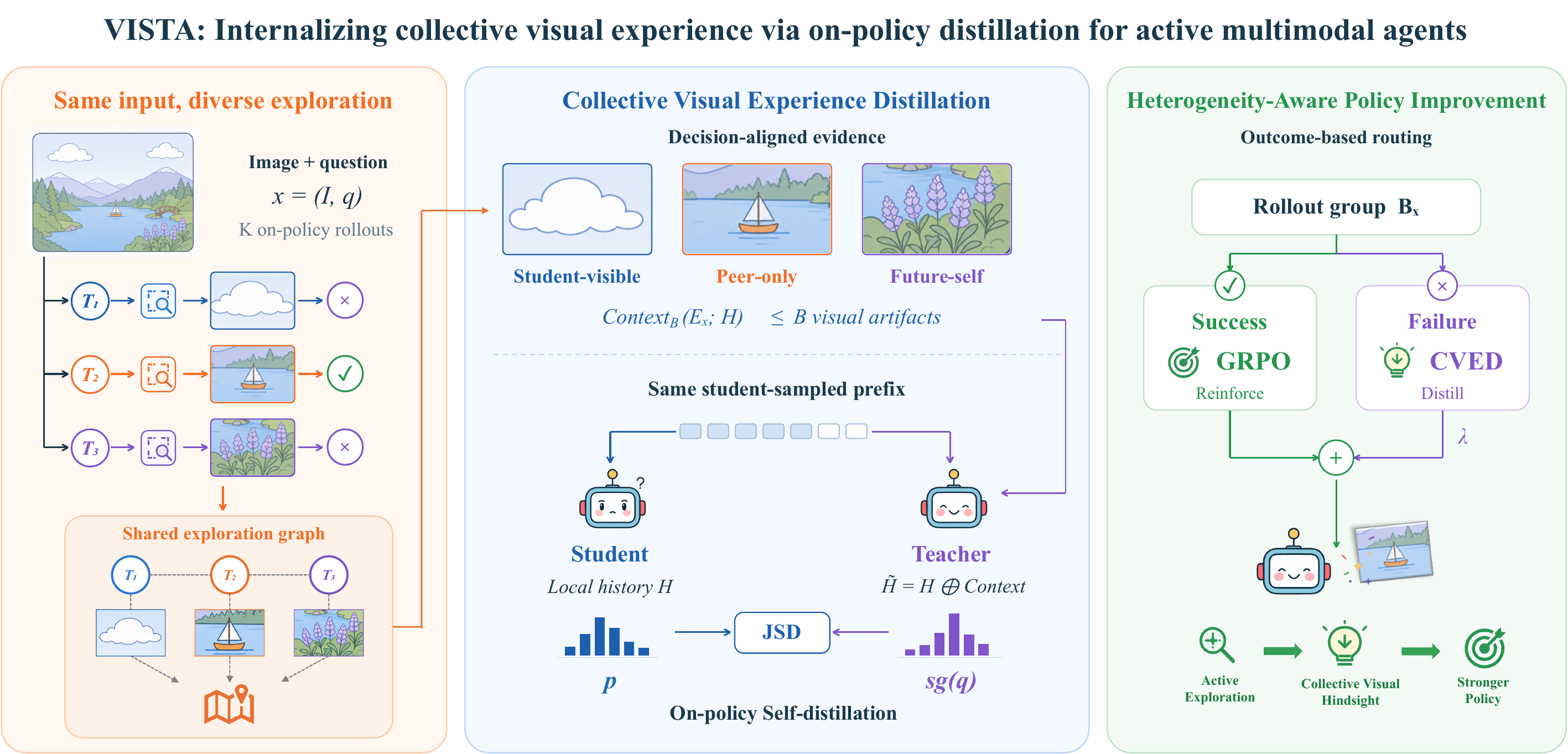}
\caption{\method combines experience transfer with heterogeneous policy improvement.
The complete rollout group supplies experience for \cved, while \hapi allocates reinforcement and distillation according to trajectory outcomes.
The teacher scores the student's sampled responses without generating new training trajectories.}
\label{fig:overview}
\vspace{-1mm}
\end{figure}

\method turns the visual experience of an on-policy rollout group into supervision for an individual agent.
Its two components address complementary aspects of this learning problem.
\cved constructs a decision-aligned teacher from the group's interaction experience, providing distributional guidance on the student's own responses.
\hapi determines how this guidance is used, reinforcing successful trajectories with RLVR and applying collective-experience distillation to unsuccessful ones.

\subsection{Problem Formulation}
\label{sec:problem}

Given an image--question pair $x=(I,q)$, an active multimodal policy alternates between generating responses and observing tool results.
At decision $t$, the interaction history $H_t$ contains the original input and all preceding responses and tool observations.
The student policy $\student$ samples a response $y_t$, which can invoke a visual tool or provide a final answer.
A tool result becomes part of the next history.
For each input, we collect
\begin{equation}
    \mathcal{B}_x=\{\traj_k\}_{k=1}^{K},
    \qquad
    \traj_k\sim\pi_{\theta_{\mathrm{old}}}(\cdot\mid x),
\end{equation}
where $\theta_{\mathrm{old}}$ denotes the policy used for rollout collection.
Each trajectory receives a verified task reward and a terminal status indicating success, an incorrect answer, a protocol-invalid action, or an incomplete attempt.
We use these outcomes to choose the learning objective, while retaining observations from the full group as potential sources of experience.
The resulting training problem is to improve the policy using what the group discovered, without changing the histories on which its responses were sampled.

\subsection{Collective Visual Experience Distillation}
\label{sec:cved}

\paragraph{Collective experience.}
A rollout group provides several attempts to acquire evidence about the same input.
Their value is not limited to the final answers.
A trajectory may reveal a small visual detail that its peers missed, or obtain a useful observation before making a subsequent reasoning error.
\cved compiles these discoveries into a task-local experience object $\experience_x=\mathcal{C}(\mathcal{B}_x)$.
It associates each visual artifact with the decision that produced it and preserves its position within the originating interaction.
Relations between artifacts retain how successive crops were derived from one another and which regions they cover.
Repeated views can share a visual representation without losing their separate trajectory occurrences.
This organization lets the teacher interpret a visual discovery in context rather than treating the group as an unordered collection of images.
Trajectory outcomes provide feedback for interpreting this experience, while both successful and unsuccessful attempts remain eligible evidence sources.

\paragraph{Decision-aligned context.}
The experience is shared across the group, but its relevance depends on the decision being supervised.
For a response generated from history $H$, the teacher is conditioned on an augmented history
\begin{equation}
    \widetilde H=H\oplus\operatorname{Context}_B(\experience_x;H),
\end{equation}
where $\operatorname{Context}_B$ selects and encodes collective experience under a budget of $B$ visual artifacts, and $\oplus$ denotes context augmentation.
Alignment distinguishes artifacts already observed by the student from those available through peers or later in its own trajectory.
Thus, the teacher can use hindsight without confusing it with evidence available when the student acted.

Collective experience therefore provides a broader evidence base for supervising decisions made under a partial interaction history.
By conditioning the teacher on this experience, \cved allows discoveries from one trajectory to inform learning in another, including useful observations from attempts that ultimately fail.
This cross-trajectory transfer expands the information available for supervision while keeping policy learning grounded in the student's own behavior.
Teacher-context construction is detailed in Appendix~\ref{sec:channels}.

\paragraph{On-policy knowledge transfer.}
Let $y$ be a student-generated response and $y_{<n}$ its prefix before token $n$.
The student $\student$ and the experience-conditioned teacher $\teacher$ evaluate this same prefix using $H$ and $\widetilde H$, respectively.
For a subset $\mathcal{S}$ of the rollout group, the on-policy distillation objective is
\begin{equation}
\label{eq:cve_loss}
    \mathcal{L}_{\mathrm{CVED}}(\theta;\mathcal{S})
    =\mathbb{E}_{(H,y_{<n})\sim\mathcal{S}}\!\left[
        D\!\left(
            \student(\cdot\mid H,y_{<n}),\,
            \teacher(\cdot\mid\widetilde H,y_{<n})
        \right)
    \right],
\end{equation}
where the expectation denotes the empirical average over generated response-token prefixes in $\mathcal{S}$, excluding prompt tokens, tool observations, and padding.
The teacher distribution is held fixed during student optimization.
Its context is drawn from the full rollout group even when only a subset receives distillation.
Thus, the teacher supplies distributional supervision on student-visited prefixes rather than generating replacement trajectories.
The additional experience changes how the teacher evaluates the student's behavior, and the resulting distributional signal trains the policy under its ordinary local history.

\subsection{Heterogeneity-Aware Policy Improvement}
\label{sec:hapi}

Trajectories within a rollout group differ in both behavioral quality and informational value.
An unsuccessful attempt may discover useful visual evidence even when its subsequent decisions lead to an incorrect answer, whereas a successful attempt provides behavior supported by a verified outcome.
These differences motivate distinct forms of policy improvement.
While \cved makes experience available across the group, \hapi determines how each trajectory learns from it.

\hapi reinforces successful trajectories through outcome-based policy optimization and applies experience-guided distillation to unsuccessful ones.
For successful trajectories, the verified reward supports strengthening behavior that already solves the task.
For unsuccessful trajectories, the experience-conditioned teacher provides distributional guidance informed by discoveries beyond the agent's local history.
This guidance covers the full sequence of generated responses, allowing collective experience to inform intermediate decisions as well as the final answer.
Both types of trajectory remain sources of collective experience, irrespective of the objective applied to their behavior.

Let $\mathcal{B}_x^{-}$ denote the unsuccessful trajectories in the group, including invalid and incomplete attempts.
Writing $\mathcal{L}_{\mathrm{GRPO}}^{\mathrm{succ}}$ for the GRPO objective restricted to successful trajectories, the overall objective is
\begin{equation}
\label{eq:joint_loss}
    \mathcal{L}_{\mathrm{VISTA}}(\theta)
    =\mathbb{E}_{x,\mathcal{B}_x}\Big[
        \mathcal{L}_{\mathrm{GRPO}}^{\mathrm{succ}}(\theta;\mathcal{B}_x)
        +\lambda\mathcal{L}_{\mathrm{CVED}}(\theta;\mathcal{B}_x^{-})
    \Big].
\end{equation}
Here $\lambda$ controls the strength of experience distillation.
Relative advantages are computed from the full group's rewards, while the CVED teacher draws on experience contributed by all trajectories.
The two objectives provide complementary directions for learning: reinforcement strengthens successful behavior, and distillation guides improvement where the current policy has not yet succeeded.
Together, \cved and \hapi couple shared experience with differentiated supervision to improve the active interaction policy.

We use Jensen--Shannon divergence for distribution matching; teacher updates and optimization details are provided in Appendix~\ref{sec:implementation}.
At inference, the student continues to use visual tools under its own interaction history, without the collective context or teacher.

\section{Experiments}
\label{sec:experiments}

Our evaluation examines the effectiveness of collective visual experience and the roles of its two components.
We first compare \method with existing agents, examine generalization across task families under different training methods, and then isolate the contributions of CVED and HAPI.
Further analyses examine teacher updates and distribution matching, generalization across backbones, and the effects of experience scale and distillation strength.

\subsection{Experimental Setup}
\label{sec:exp_setup}

\paragraph{Models and training.}
We use Qwen3-VL-4B-Instruct and Qwen3-VL-8B-Instruct \citep{bai2025qwen3vl} as the default backbones and train on the Vision-OPD-6K split \citep{yuan2026visionopd}.
Training inputs contain the original full image without target-box overlays and the question without added spatial-restriction hints; the student acquires additional observations through visual tools.
Our default configuration uses $K=8$ rollouts per input, distillation weight $\lambda=0.1$, and an artifact budget of $B=4$ for teacher-context construction.
Answer correctness provides the task reward, and the policy may answer directly without invoking a tool.

\paragraph{Benchmarks and metrics.}
The main evaluation covers V*Bench \citep{wu2023vstar}, ZoomBench \citep{wei2026zooming}, HR-Bench-4K/8K \citep{wang2024dc2}, and the English and Chinese splits of MME-RealWorld \citep{zhang2024mmerealworld}.
To assess whether improvements extend beyond fine-grained perception, the training-method comparison also includes MMStar, MMBench, and MathVista \citep{chen2024we,liu2024mmbench,lu2024mathvista}.
Component ablations and parameter analyses use V*Bench and both HR-Bench variants unless otherwise specified.
We report accuracy (\%) and the unweighted mean over the benchmarks in each table. Bold and underlined scores indicate the best and second-best results, respectively.

\paragraph{Baselines and comparison protocol.}
We compare against the base checkpoint, GRPO \citep{shao2024deepseekmath}, and OPSD, VCSD, Vision-OPD, and Imagine-OPD \citep{zhao2026opsd,liang2026vcsd,yuan2026visionopd,cai2026imagineopd}.
Within each backbone, controlled comparisons share the starting checkpoint, training data, optimizer and schedule, with matched visual tools and interaction limits.
Evaluation is deterministic, and all experimental results are averaged over three seeds.
All reported results are obtained from our own experiments, including the broader comparison with thinking-with-images agents and general multimodal models; larger and proprietary models serve as capability references.
Appendix~\ref{sec:additional_exp_setup} provides the reference-model descriptions and further protocol details.

\FloatBarrier
\subsection{Main Results}
\label{sec:main_results}

\paragraph{Consistent improvements across visual tasks.}
\method improves on its base checkpoints across all six benchmarks at both scales (Table~\ref{tab:main_results}), with the largest gains on ZoomBench.
The improvements on both HR-Bench variants and both language splits of MME-RealWorld show that the benefit extends beyond a single visual-search benchmark.
Both variants also outperform the compared thinking-with-images baselines throughout the suite.

The trained 4B model exceeds the base 8B checkpoint across this comparison, suggesting that better use of interaction experience can complement increases in model capacity.
Although the trained 8B model generally performs better, the consistent gains at both scales support collective experience as a useful source of supervision rather than a capacity-specific remedy.

\begin{table}[!htb]
\vspace{-1mm}
\centering
\caption{Performance comparison on visual perception and reasoning benchmarks. We report accuracy (\%). $\Delta$ denotes the gain over the corresponding Qwen3-VL base model.}
\label{tab:main_results}
\footnotesize
\setlength{\tabcolsep}{2.4pt}
\renewcommand{\arraystretch}{1.12}
\resizebox{\linewidth}{!}{%
\begin{tabular}{@{}l*{7}{c}@{}}
\toprule
\textbf{Model} & \textbf{V* Bench} & \textbf{ZoomBench} & \textbf{HR-Bench-4K} & \textbf{HR-Bench-8K} & \textbf{MME-RW-EN} & \textbf{MME-RW-CN} & \textbf{Average} \\
\midrule
\rowcolor{vistaClosed}\multicolumn{8}{c}{\strut\textbf{Closed-source models}} \\
GPT-5.2 & 79.06 & 50.89 & 81.12 & 78.38 & \underline{72.60} & 68.80 & 71.81 \\
GPT-5.4 & 76.96 & 52.66 & \textbf{84.00} & 77.88 & \textbf{74.20} & \textbf{70.93} & 72.77 \\
Gemini-2.5-Pro & 78.01 & 49.82 & 80.63 & 77.50 & 71.29 & 69.11 & 71.06 \\
Gemini-3-Flash & 78.53 & 50.06 & 80.87 & 78.00 & 71.35 & 69.38 &  71.37 \\
\midrule
\rowcolor{vistaOpen}\multicolumn{8}{c}{\strut\textbf{Open-source models}} \\
Qwen3-VL-4B & 81.68 & 44.97 & 78.50 & 76.25 & 63.27 & 62.92 & 67.93 \\
Qwen3-VL-8B & 84.82 & 42.96 & 79.63 & 75.25 & 63.19 & 64.61 & 68.41 \\
MiMo-VL-RL-7B & 83.25 & 45.68 & 73.50 & 69.38 & 62.73 & 55.89 & 65.07 \\
MiniCPM-V-4.5-9B & 70.68 & 42.60 & 69.63 & 61.50 & 62.65 & 61.64 & 61.45 \\
GLM-4.6V-106B & 86.91 & 50.06 & 82.13 & 78.88 & 65.57 & 65.62 & 71.53 \\
Kimi-K2.6-1T & \underline{88.48} & 53.14 & 81.88 & 78.00 & 69.22 & 66.13 & 72.81 \\
\midrule
\rowcolor{vistaAgent}\multicolumn{8}{c}{\strut\textbf{Thinking-with-images models}} \\
Thyme-7B & 82.20 & 45.09 & 77.00 & 72.00 & 64.80 & 64.59 & 67.61 \\
DeepEyes-7B & 85.86 & 46.51 & 75.13 & 72.63 & 64.10 & 64.09 & 68.05 \\
DeepEyesV2-7B & 81.68 & 44.97 & 77.88 & 73.75 & 64.90 & 65.07 & 68.04 \\
Pixel-Reasoner-7B & 84.29 & 44.73 & 76.50 & 72.00 & 64.40 & 64.12 & 67.67 \\
\midrule
\rowcolor{vistaOurs}\multicolumn{8}{c}{\strut\textbf{Our models}} \\
\rowcolor{vistaHighlight}
\textbf{\method{}-4B} & \underline{88.48} & \underline{53.85} & 83.50 & \underline{79.50} & 68.24 & 69.29 & \underline{73.81} \\
\quad $\Delta$ (vs. Qwen3-VL-4B) & +6.80 & +8.88 & +5.00 & +3.25 & +4.97 & +6.37 & +5.88 \\
\rowcolor{vistaHighlight}
\textbf{\method{}-8B} & \textbf{89.53} & \textbf{54.43} & \underline{83.75} & \textbf{79.88} & 68.19 & \underline{69.55} & \textbf{74.22} \\
\quad $\Delta$ (vs. Qwen3-VL-8B) & +4.71 & +11.47 & +4.12 & +4.63 & +5.00 & +4.94 & +5.81 \\
\bottomrule
\end{tabular}}
\vspace{-1mm}
\end{table}

\FloatBarrier
\subsection{Cross-Task Generalization}
\label{sec:cross_task_generalization}

\textbf{Generalization beyond fine-grained perception.}
Table~\ref{tab:training_comparison} pairs fine-grained visual tasks with MMStar, MMBench, and MathVista to examine the breadth of the learned capabilities under fixed backbones.
\method leads both task groups at 4B and 8B, with improvements in fine-grained perception accompanied by stronger general reasoning rather than a trade-off between them.
The result on MathVista further extends this pattern to mathematical reasoning in visual contexts.

\textbf{Consistent gains across training methods and scales.}
GRPO and the competing distillation methods also improve over the base checkpoints, but their relative strengths vary across tasks.
In contrast, \method maintains its advantage on every evaluated benchmark at both scales.
This consistency supports cross-task generalization within the evaluated suite, rather than an average gain driven by one task family.
The ablations below distinguish the contributions of experience construction and supervision allocation.

\begin{table}[t]
\vspace{-1mm}
\centering
\caption{Comparison of training methods on fine-grained visual tasks and general reasoning tasks. We report accuracy (\%) for Qwen3-VL-4B and Qwen3-VL-8B.}
\label{tab:training_comparison}
\fontsize{7.71}{9.64}\selectfont
\setlength{\tabcolsep}{2.4pt}
\renewcommand{\arraystretch}{1.12}
\begin{tabularx}{\linewidth}{l>{\centering\arraybackslash}Xcc*{4}{>{\centering\arraybackslash}X}}
\toprule
\multirow{2}{*}{\textbf{Methods}} & \multicolumn{3}{c}{\textbf{Fine-Grained Visual Tasks}} & \multicolumn{3}{c}{\textbf{General Reasoning Tasks}} & \multirow{2}{*}{\textbf{Average}} \\
\cmidrule(lr){2-4}\cmidrule(lr){5-7}
& \textbf{V* Bench} & \textbf{HR-Bench-4K} & \textbf{HR-Bench-8K} & \textbf{MMStar} & \textbf{MMBench} & \textbf{MathVista} & \\
\midrule
\rowcolor{vistaOpen}\multicolumn{8}{c}{\strut\textbf{Qwen3-VL-4B}} \\
Base & 81.68 & 78.50 & 76.25 & 68.73 & 81.22 & 73.70 & 76.68 \\
GRPO & 84.12 & 80.08 & 77.50 & 70.87 & 82.48 & 75.10 & 78.36 \\
OPSD & 85.34 & 81.25 & \underline{78.25} & \underline{71.53} & 82.91 & 75.60 & 79.15 \\
VCSD & 83.77 & 80.38 & 76.88 & 69.20 & \underline{83.08} & 74.55 & 77.98 \\
Vision-OPD & 85.34 & 81.25 & 77.38 & 70.55 & 82.72 & 75.65 & 78.82 \\
Imagine-OPD & \underline{86.39} & \underline{82.00} & 77.80 & 71.47 & 82.95 & \underline{75.90} & \underline{79.42} \\
\rowcolor{vistaHighlight}\textbf{\method} & \textbf{88.48} & \textbf{83.50} & \textbf{79.50} & \textbf{73.07} & \textbf{83.52} & \textbf{76.30} & \textbf{80.73} \\
\midrule
\rowcolor{vistaAgent}\multicolumn{8}{c}{\strut\textbf{Qwen3-VL-8B}} \\
Base & 84.82 & 79.63 & 75.25 & 70.67 & 82.68 & 76.50 & 78.26 \\
GRPO & 86.91 & 82.13 & 78.25 & 73.07 & 83.62 & 77.80 & 80.30 \\
OPSD & 87.96 & 82.75 & 79.00 & 74.20 & 83.91 & 78.10 & 80.99 \\
VCSD & 87.43 & \underline{83.25} & \underline{79.13} & 74.35 & 83.18 & 77.65 & 80.83 \\
Vision-OPD & \underline{88.48} & 82.88 & 78.63 & 74.65 & 83.82 & 78.35 & 81.14 \\
Imagine-OPD & 87.96 & 83.05 & 78.88 & \underline{75.15} & \underline{84.12} & \underline{78.50} & \underline{81.28} \\
\rowcolor{vistaHighlight}\textbf{\method} & \textbf{89.53} & \textbf{83.75} & \textbf{79.88} & \textbf{76.67} & \textbf{84.36} & \textbf{78.80} & \textbf{82.17} \\
\bottomrule
\end{tabularx}
\vspace{-1mm}
\end{table}

\subsection{Ablation Study}
\label{sec:ablation}

We isolate experience construction and supervision allocation on Qwen3-VL-4B.
The CVED ablation fixes HAPI and varies the teacher's privileged context, whereas the HAPI ablation fixes CVED and varies the learning rule.

\paragraph{CVED: the content and organization of experience.}
\label{sec:cved_ablation}

Table~\ref{tab:cved_ablation} compares CVED with reference-answer supervision (Answer-only), annotated evidence crops (Oracle ROI), and trajectory-ordered observations and interaction records (Unstructured).
All variants use the same training set, and CVED and Unstructured share the selected observations and context budget.
CVED performs best across all three tasks, followed by Oracle ROI, Unstructured, and Answer-only.

The ordering suggests that the teacher benefits from visual evidence beyond knowing the correct endpoint, while the strength of Oracle ROI highlights the value of relevant, localized information.
CVED improves further by retaining interaction context and aligning evidence with the student's decision.
Its advantage over Unstructured is particularly informative because the observations and budget are unchanged: organizing the evidence, rather than simply supplying more images, makes the supervision more effective.
Together with the Oracle ROI comparison, this suggests that localization and interaction context serve complementary roles: useful supervision depends on both the evidence presented and its relation to the decision being learned.

\paragraph{HAPI: allocating complementary learning signals.}
\label{sec:hapi_ablation}

Table~\ref{tab:hapi_ablation} compares GRPO-only, CVED-only, Uniform hybrid, and HAPI with the same rollout-group size and optimization budget.
Uniform hybrid applies both objectives to every trajectory, whereas HAPI reinforces successful trajectories and distills unsuccessful ones.
All distillation variants share the same CVED context construction.

Uniform hybrid outperforms either objective alone, and HAPI improves further across all three tasks.
Thus, reinforcement and distillation are complementary, but their allocation also matters.
Since Uniform hybrid already receives both signals, HAPI's advantage supports matching supervision to trajectory outcomes rather than merely adding a loss term.
Meanwhile, unsuccessful attempts can still contribute useful observations to CVED, separating their informational value from how their behavior is optimized.
Thus, the routing rule changes how trajectories learn without restricting the shared evidence pool to successful attempts, preserving potentially useful discoveries from failures.

\begin{table}[!htbp]
\vspace{-1mm}
\centering
\begin{minipage}[t]{0.49\linewidth}
\vspace{0pt}
\caption{CVED ablation: privileged supervision under fixed HAPI routing.}
\label{tab:cved_ablation}
\centering
\fontsize{7.71}{9.64}\selectfont
\setlength{\tabcolsep}{0.7pt}
\renewcommand{\arraystretch}{1.12}
\begin{tabularx}{\linewidth}{@{}l*{4}{>{\centering\arraybackslash}X}@{}}
\toprule
Setting & V* & HR-4K & HR-8K & Avg. \\
\midrule
Answer-only & 86.56 & 81.30 & 77.80 & 81.89 \\
Unstructured & 86.91 & 82.13 & 78.25 & 82.43 \\
Oracle ROI & \underline{87.96} & \underline{82.80} & \underline{79.00} & \underline{83.25} \\
\rowcolor{vistaHighlight}\textbf{CVED} & \textbf{88.48} & \textbf{83.50} & \textbf{79.50} & \textbf{83.83} \\
\bottomrule
\end{tabularx}
\end{minipage}\hfill
\begin{minipage}[t]{0.49\linewidth}
\vspace{0pt}
\caption{HAPI ablation: learning rules under fixed experience construction.}
\label{tab:hapi_ablation}
\centering
\fontsize{7.71}{9.64}\selectfont
\setlength{\tabcolsep}{0.7pt}
\renewcommand{\arraystretch}{1.12}
\begin{tabularx}{\linewidth}{@{}l*{4}{>{\centering\arraybackslash}X}@{}}
\toprule
Setting & V* & HR-4K & HR-8K & Avg. \\
\midrule
GRPO-only & 84.12 & 80.08 & 77.50 & 80.57 \\
CVED-only & 86.21 & 81.50 & 78.13 & 81.95 \\
Uniform hybrid & \underline{87.43} & \underline{82.38} & \underline{78.80} & \underline{82.87} \\
\rowcolor{vistaHighlight}\textbf{HAPI} & \textbf{88.48} & \textbf{83.50} & \textbf{79.50} & \textbf{83.83} \\
\bottomrule
\end{tabularx}
\end{minipage}
\vspace{-1mm}
\end{table}

\subsection{Training Design Analysis}
\label{sec:training_design}

We next examine two supporting choices within the fixed CVED--HAPI framework: how the teacher is updated and how its distributions are matched.
Both comparisons use Qwen3-VL-4B and the six-benchmark suite from Table~\ref{tab:main_results}.

\paragraph{Teacher update strategy.}
Freezing the teacher at initialization retains much of the performance of EMA (Table~\ref{tab:teacher_update}).
EMA gives the stronger average, especially on V*Bench and ZoomBench, while the frozen teacher performs better on HR-Bench-8K and MME-RW-CN.
We therefore adopt EMA as an aggregate preference rather than a uniformly better choice for every task.

Both teachers receive newly collected experience, so fixed weights do not imply a fixed teaching signal.
The competitive frozen-teacher result suggests that context enrichment can provide useful supervision in its own right, with parameter updates offering an additional, task-dependent benefit.

\begin{table}[!htbp]
\vspace{-1mm}
\centering
\caption{Teacher-update ablation on Qwen3-VL-4B.}
\label{tab:teacher_update}
\fontsize{7.71}{9.64}\selectfont
\setlength{\tabcolsep}{2.4pt}
\renewcommand{\arraystretch}{1.12}
\begin{tabularx}{\linewidth}{@{}>{\raggedright\arraybackslash}X*{5}{>{\centering\arraybackslash}X}>{\hsize=1.1\hsize\linewidth=\hsize\centering\arraybackslash}X>{\hsize=0.9\hsize\linewidth=\hsize\centering\arraybackslash}X@{}}
\toprule
Teacher & V*Bench & ZoomBench & HR-Bench-4K & HR-Bench-8K & MME-RW-EN & MME-RW-CN & Avg. \\
\midrule
Frozen initial & \underline{87.43} & \underline{51.95} & \underline{83.38} & \textbf{80.75} & \underline{68.19} & \textbf{69.51} & \underline{73.54} \\
\rowcolor{vistaHighlight}EMA (ours) & \textbf{88.48} & \textbf{53.85} & \textbf{83.50} & \underline{79.50} & \textbf{68.24} & \underline{69.29} & \textbf{73.81} \\
\bottomrule
\end{tabularx}
\vspace{-1mm}
\end{table}

\paragraph{Divergence objective.}
\label{sec:divergence_analysis}

Table~\ref{tab:divergence_objective} compares forward KL, $D_{\mathrm{KL}}(\teacher\|\student)$, reverse KL, $D_{\mathrm{KL}}(\student\|\teacher)$, and JSD with an equally weighted mixture.
All variants match distributions on student-generated prefixes with the same CVED context construction, HAPI routing, and teacher updates.

JSD and reverse KL outperform forward KL across the suite.
JSD gives the best average, while reverse KL is stronger on ZoomBench and HR-Bench-8K.
Since all three variants improve over the base model, collective-experience learning is not tied to a particular divergence.
The task-dependent ordering supports JSD as an effective overall choice without implying that it is optimal for every benchmark.

\begin{table}[!htbp]
\vspace{-1mm}
\centering
\caption{Divergence-objective comparison on Qwen3-VL-4B with fixed CVED and HAPI.}
\label{tab:divergence_objective}
\fontsize{7.71}{9.64}\selectfont
\setlength{\tabcolsep}{2.4pt}
\renewcommand{\arraystretch}{1.12}
\begin{tabularx}{\linewidth}{@{}>{\raggedright\arraybackslash}X*{5}{>{\centering\arraybackslash}X}>{\hsize=1.1\hsize\linewidth=\hsize\centering\arraybackslash}X>{\hsize=0.9\hsize\linewidth=\hsize\centering\arraybackslash}X@{}}
\toprule
Objective & V*Bench & ZoomBench & HR-Bench-4K & HR-Bench-8K & MME-RW-EN & MME-RW-CN & Avg. \\
\midrule
Forward KL & 86.91 & 50.89 & 82.25 & 77.80 & 67.10 & 68.10 & 72.18 \\
Reverse KL & \underline{87.96} & \textbf{53.96} & \underline{82.95} & \textbf{79.75} & \underline{67.80} & \underline{69.05} & \underline{73.58} \\
\rowcolor{vistaHighlight}\textbf{JSD} & \textbf{88.48} & \underline{53.85} & \textbf{83.50} & \underline{79.50} & \textbf{68.24} & \textbf{69.29} & \textbf{73.81} \\
\bottomrule
\end{tabularx}
\vspace{-1mm}
\end{table}

\subsection{Backbone Generalization}
\label{sec:backbone_generalization}

We repeat the base, GRPO, OPSD, and \method comparisons on Qwen3.5-4B and Qwen3.5-9B \citep{qwen2026qwen35} using the same protocol within each backbone (Table~\ref{tab:backbone_generalization}).
\method leads on all three tasks for both backbones.
Its advantage over OPSD is concentrated on V*Bench at 4B, where the high-resolution results are close, but extends more clearly across the suite at 9B.

On Qwen3.5-9B, OPSD is stronger than GRPO on V*Bench but weaker on the high-resolution tasks, whereas \method improves over both throughout.
The combined supervision therefore remains useful across the tested backbones and across tasks that favor different baseline objectives, although the size of the benefit depends on the model and task.

\begin{table}[!htbp]
\vspace{-1mm}
\centering
\caption{Backbone generalization on Qwen3.5-4B and Qwen3.5-9B. All methods use the same training and evaluation protocol within each backbone.}
\label{tab:backbone_generalization}
\begin{minipage}[t]{0.49\linewidth}
\centering
\fontsize{7.4}{8.9}\selectfont
\textbf{Qwen3.5-4B}\\[2pt]
\setlength{\tabcolsep}{2pt}
\renewcommand{\arraystretch}{1.1}
\begin{tabularx}{\linewidth}{@{}>{\raggedright\arraybackslash}X*{4}{>{\centering\arraybackslash}X}@{}}
\toprule
Method & V* & HR-4K & HR-8K & Avg. \\
\midrule
Base & 82.20 & 82.75 & 78.38 & 81.11 \\
GRPO & 83.25 & 82.88 & 78.75 & 81.63 \\
OPSD & \underline{84.29} & \underline{83.00} & \underline{79.00} & \underline{82.10} \\
\rowcolor{vistaHighlight}\textbf{\method} & \textbf{88.48} & \textbf{83.13} & \textbf{79.13} & \textbf{83.58} \\
\bottomrule
\end{tabularx}
\end{minipage}\hfill
\begin{minipage}[t]{0.49\linewidth}
\centering
\fontsize{7.4}{8.9}\selectfont
\textbf{Qwen3.5-9B}\\[2pt]
\setlength{\tabcolsep}{2pt}
\renewcommand{\arraystretch}{1.1}
\begin{tabularx}{\linewidth}{@{}>{\raggedright\arraybackslash}X*{4}{>{\centering\arraybackslash}X}@{}}
\toprule
Method & V* & HR-4K & HR-8K & Avg. \\
\midrule
Base & 81.68 & 84.50 & 79.88 & 82.02 \\
GRPO & 84.29 & \underline{85.75} & \underline{82.38} & 84.14 \\
OPSD & \underline{87.96} & 84.25 & 81.25 & \underline{84.49} \\
\rowcolor{vistaHighlight}\textbf{\method} & \textbf{92.15} & \textbf{87.25} & \textbf{84.38} & \textbf{87.93} \\
\bottomrule
\end{tabularx}
\end{minipage}
\vspace{-1mm}
\end{table}

\subsection{Scalability}
\label{sec:scalability}

Increasing the rollout-group size $K\in\{4,6,8,12,16\}$ improves or preserves accuracy on all three benchmarks, with the strongest response on V*Bench (Figure~\ref{fig:sensitivity}(a)).
Gains continue beyond the default group size but diminish toward the largest tested groups, indicating that additional attempts remain useful with decreasing returns.

Because the artifact budget is fixed, larger groups expand the candidate experience pool without requiring the teacher to process every new observation.
Together with the budget sweep below, this favors collecting broadly and conditioning selectively.
The sweep also increases rollout volume, so it measures scaling with additional experience rather than a fixed-compute allocation.
In particular, the different trends for $K$ and $B$ suggest that the supply of candidate evidence and the teacher's capacity to use it should be treated as separate design choices.

\subsection{Hyperparameter Sensitivity}
\label{sec:hyperparam}

Having varied the experience pool, we examine how strongly that experience should supervise the student and how much should enter each teacher context.
We vary $\lambda$ and $B$ independently on Qwen3-VL-4B, keeping the other settings at their defaults.

\begin{figure}[!htbp]
\vspace{-1mm}
\centering
\includegraphics[width=\linewidth]{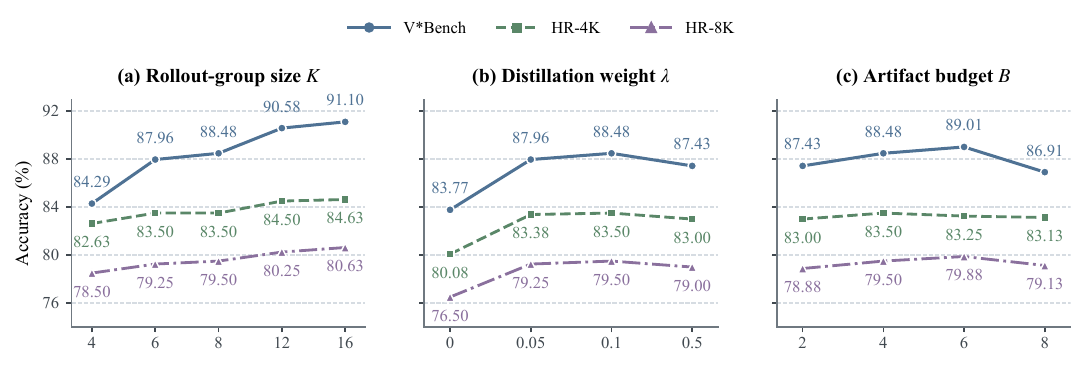}
\caption{Sensitivity of VISTA on Qwen3-VL-4B to (a) rollout-group size $K$, (b) distillation weight $\lambda$, and (c) visual artifact budget $B$. Points report V*Bench, HR-4K, and HR-8K accuracy. Tested settings are equally spaced along each horizontal axis.}
\label{fig:sensitivity}
\vspace{-1mm}
\end{figure}

\paragraph{Distillation strength.}
Figure~\ref{fig:sensitivity}(b) varies $\lambda\in\{0,0.05,0.1,0.5\}$ under fixed HAPI routing; setting $\lambda=0$ retains only successful-trajectory reinforcement in Equation~\ref{eq:joint_loss}.
Every tested nonzero weight improves all three tasks over this zero-distillation setting, while the shared peak at $\lambda=0.1$ favors a moderate distillation strength.
The decline at the largest weight shows that increasing emphasis on the teacher signal is not uniformly beneficial, although the gains across several nonzero settings indicate that the benefit is not confined to one coefficient.
The common peak across tasks also supports a shared default rather than selecting a separate weight for each benchmark.

\paragraph{Visual artifact budget.}
Figure~\ref{fig:sensitivity}(c) shows that an intermediate budget is preferable to the smallest or largest tested context: HR-4K favors $B=4$, while V*Bench and HR-8K favor $B=6$.
Longer teacher contexts may make decision-relevant evidence harder to integrate, while processing additional artifacts increases computation and can prolong training.
We retain $B=4$ as a compact default.
In contrast to the continued gains from larger rollout groups, this non-monotonic trend supports expanding the experience pool while selecting a concise context for each decision.

\section{Conclusion}
\label{sec:conclusion}

We reframed grouped active-agent training as an experience-internalization problem.
Instead of using on-policy rollouts only to estimate scalar advantages, \method constructs collective visual experience through \cved and uses \hapi to combine successful-behavior reinforcement with experience-guided distillation on unsuccessful trajectories.
The deployed agent remains a single ordinary tool-using policy.
This view makes experience sources, interaction structure, and outcome feedback independently testable components of privileged supervision.
More broadly, it suggests that the transient population created by on-policy training can serve as a renewable source of experience for the policy itself.
Appendix~\ref{sec:limitations} discusses limitations and directions for future work.

\label{sec:main_text_end}
\subsection*{AI use statement}

Large language models were used solely to assist with language polishing, including correcting grammar, refining phrasing, and improving the clarity and readability of the manuscript.
The authors reviewed and revised the suggested edits to ensure that they accurately conveyed the intended technical meaning.
All experimental results and implementation details reported in the paper were checked and confirmed by the authors.
The authors take full responsibility for the accuracy and integrity of the final manuscript, including its technical claims and conclusions.

\subsection*{Reproducibility statement}

Section~\ref{sec:method} specifies the learning problem, \cved, \hapi, and the training objective. Appendix~\ref{sec:implementation} describes experience construction and teacher supervision.
Section~\ref{sec:experiments} describes the evaluation protocol and experimental comparisons.
Appendix~\ref{sec:prompt_templates} provides the student and teacher prompt templates, and Appendix~\ref{sec:channels} specifies the information supplied to the teacher.
Code is available at \url{https://github.com/jiangz20/VISTA}.

\bibliography{references}

@article{wu2023vstar,
  title         = {{V*: Guided Visual Search as a Core Mechanism in Multimodal LLMs}},
  author        = {Wu, Penghao and Xie, Saining},
  journal       = {arXiv preprint arXiv:2312.14135},
  year          = {2023},
  eprint        = {2312.14135},
  archiveprefix = {arXiv},
  primaryclass  = {cs.CV}
}

@article{hu2024visualsketchpad,
  title         = {{Visual Sketchpad: Sketching as a Visual Chain of Thought for Multimodal Language Models}},
  author        = {Hu, Yushi and Shi, Weijia and Fu, Xingyu and Roth, Dan and Ostendorf, Mari and Zettlemoyer, Luke and Smith, Noah A. and Krishna, Ranjay},
  journal       = {arXiv preprint arXiv:2406.09403},
  year          = {2024},
  eprint        = {2406.09403},
  archiveprefix = {arXiv},
  primaryclass  = {cs.CV}
}

@article{shen2024zoomeye,
  title         = {{ZoomEye: Enhancing Multimodal LLMs with Human-Like Zooming Capabilities through Tree-Based Image Exploration}},
  author        = {Shen, Haozhan and Zhao, Kangjia and Zhao, Tiancheng and Xu, Ruochen and Zhang, Zilun and Zhu, Mingwei and Yin, Jianwei},
  journal       = {arXiv preprint arXiv:2411.16044},
  year          = {2024},
  eprint        = {2411.16044},
  archiveprefix = {arXiv},
  primaryclass  = {cs.CV}
}

@article{zheng2025deepeyes,
  title         = {{DeepEyes: Incentivizing ``Thinking with Images'' via Reinforcement Learning}},
  author        = {Zheng, Ziwei and Yang, Michael and Hong, Jack and Zhao, Chenxiao and Xu, Guohai and Yang, Le and Shen, Chao and Yu, Xing},
  journal       = {arXiv preprint arXiv:2505.14362},
  year          = {2025},
  eprint        = {2505.14362},
  archiveprefix = {arXiv},
  primaryclass  = {cs.CV}
}

@article{wu2025vtoolr1,
  title         = {{VTool-R1: VLMs Learn to Think with Images via Reinforcement Learning on Multimodal Tool Use}},
  author        = {Wu, Mingyuan and Yang, Jingcheng and Jiang, Jize and Li, Meitang and Yan, Kaizhuo and Yu, Hanchao and Zhang, Minjia and Zhai, Chengxiang and Nahrstedt, Klara},
  journal       = {arXiv preprint arXiv:2505.19255},
  year          = {2025},
  eprint        = {2505.19255},
  archiveprefix = {arXiv},
  primaryclass  = {cs.CV}
}

@article{su2025pixelreasoner,
  title         = {{Pixel Reasoner: Incentivizing Pixel-Space Reasoning with Curiosity-Driven Reinforcement Learning}},
  author        = {Wang, Haozhe and Su, Alex and Ren, Weiming and Lin, Fangzhen and Chen, Wenhu},
  journal       = {arXiv preprint arXiv:2505.15966},
  year          = {2025},
  eprint        = {2505.15966},
  archiveprefix = {arXiv},
  primaryclass  = {cs.CV}
}

@article{liu2025visualarft,
  title         = {{Visual Agentic Reinforcement Fine-Tuning}},
  author        = {Liu, Ziyu and Zang, Yuhang and Zou, Yushan and Liang, Zijian and Dong, Xiaoyi and Cao, Yuhang and Duan, Haodong and Lin, Dahua and Wang, Jiaqi},
  journal       = {arXiv preprint arXiv:2505.14246},
  year          = {2025},
  eprint        = {2505.14246},
  archiveprefix = {arXiv},
  primaryclass  = {cs.CV}
}

@article{yang2026mapo,
  title         = {{Walk the Talk: Bridging the Reasoning-Action Gap for Thinking with Images via Multimodal Agentic Policy Optimization}},
  author        = {Yang, Wenhao and Xia, Yu and Huang, Jinlong and Lu, Shiyin and Chen, Qing-Guo and Xu, Zhao and Luo, Weihua and Zhang, Kaifu and Zhou, Yuchen and Xia, Xiaobo and Wan, Yuanyu and Zhang, Lijun and Chua, Tat-Seng},
  journal       = {arXiv preprint arXiv:2604.06777},
  year          = {2026},
  eprint        = {2604.06777},
  archiveprefix = {arXiv},
  primaryclass  = {cs.CV}
}

@article{penaloza2026privileged,
  title         = {{Privileged Information Distillation for Language Models}},
  author        = {Penaloza, Emiliano and Vattikonda, Dheeraj and Gontier, Nicolas and Lacoste, Alexandre and Charlin, Laurent and Caccia, Massimo},
  journal       = {arXiv preprint arXiv:2602.04942},
  year          = {2026},
  eprint        = {2602.04942},
  archiveprefix = {arXiv},
  primaryclass  = {cs.CL}
}

@article{ye2026opcd,
  title         = {{On-Policy Context Distillation for Language Models}},
  author        = {Ye, Tianzhu and Dong, Li and Wu, Xun and Huang, Shaohan and Wei, Furu},
  journal       = {arXiv preprint arXiv:2602.12275},
  year          = {2026},
  eprint        = {2602.12275},
  archiveprefix = {arXiv},
  primaryclass  = {cs.CL}
}

@article{yang2026opid,
  title         = {{{OPID}: On-Policy Skill Distillation for Agentic Reinforcement Learning}},
  author        = {Yang, Shuo and Wu, Jinyang and Lu, Zhengxi and Shen, Yuhao and Zhang, Fan and Feng, Lang and Zhang, Shuai and Luo, Haoran and Lian, Zheng and Wen, Zhengqi and Tao, Jianhua},
  journal       = {arXiv preprint arXiv:2606.26790},
  year          = {2026},
  eprint        = {2606.26790},
  archiveprefix = {arXiv},
  primaryclass  = {cs.LG}
}

@article{yuan2026visionopd,
  title         = {{Vision-OPD: Learning to See Fine Details for Multimodal LLMs via On-Policy Self-Distillation}},
  author        = {Yuan, Qianhao and Lou, Jie and Yu, Xing and Lin, Hongyu and Sun, Le and Han, Xianpei and Lu, Yaojie},
  journal       = {arXiv preprint arXiv:2605.18740},
  year          = {2026},
  eprint        = {2605.18740},
  archiveprefix = {arXiv},
  primaryclass  = {cs.CV}
}

@article{liu2026vaopd,
  title         = {{Visual-Advantage On-Policy Distillation for Vision-Language Models}},
  author        = {Liu, Ruiqi and Lv, Xiaolei and Li, Gengsheng and Zhu, Ximo and Wang, Zhiheng and Zhang, Zhengbo and Chen, Junkai and Li, Zhiheng and Li, Bo and Gao, Jun and Wu, Shu},
  journal       = {arXiv preprint arXiv:2605.21924},
  year          = {2026},
  eprint        = {2605.21924},
  archiveprefix = {arXiv},
  primaryclass  = {cs.CV}
}

@article{shao2024deepseekmath,
  title         = {{DeepSeekMath: Pushing the Limits of Mathematical Reasoning in Open Language Models}},
  author        = {Shao, Zhihong and Wang, Peiyi and Zhu, Qihao and Xu, Runxin and Song, Junxiao and Bi, Xiao and Zhang, Haowei and Zhang, Mingchuan and Li, Y. K. and Wu, Y. and Guo, Daya},
  journal       = {arXiv preprint arXiv:2402.03300},
  year          = {2024},
  eprint        = {2402.03300},
  archiveprefix = {arXiv},
  primaryclass  = {cs.CL}
}

@inproceedings{tarvainen2017mean,
  title     = {{Mean Teachers Are Better Role Models: Weight-Averaged Consistency Targets Improve Semi-Supervised Deep Learning Results}},
  author    = {Tarvainen, Antti and Valpola, Harri},
  booktitle = {Advances in Neural Information Processing Systems},
  volume    = {30},
  year      = {2017}
}

@article{lin1991divergence,
  title   = {{Divergence Measures Based on the Shannon Entropy}},
  author  = {Lin, Jianhua},
  journal = {IEEE Transactions on Information Theory},
  volume  = {37},
  number  = {1},
  pages   = {145--151},
  year    = {1991},
  doi     = {10.1109/18.61115}
}

@article{sun2026vzero,
  title         = {{V-Zero: Answer-Label-Free On-Policy Distillation with Contrastive Evidence Gating for Fine-Grained Visual Reasoning}},
  author        = {Sun, Haoxiang and Yi, Zhihang and Deng, Langxuan and Zhou, Yuhao and Jia, Peiqi and Zhao, Jian and Yuan, Li and Lv, Jiancheng and Wang, Tao},
  journal       = {arXiv preprint arXiv:2606.25319},
  year          = {2026},
  eprint        = {2606.25319},
  archiveprefix = {arXiv},
  primaryclass  = {cs.CV}
}

@article{li2026visualopsd,
  title         = {{Visual-OPSD: Cross-Modal On-Policy Self-Distillation for Efficient Unified Multimodal Reasoning}},
  author        = {Li, Pengyu and Gao, Zhitao and Zhang, Lingling and Huang, Muye and Li, Yuanming and Yang, Zesheng and Xu, Fangzhi and Liu, Jun},
  journal       = {arXiv preprint arXiv:2606.18974},
  year          = {2026},
  eprint        = {2606.18974},
  archiveprefix = {arXiv},
  primaryclass  = {cs.CV}
}

@article{bai2025qwen3vl,
  title         = {{{Qwen3-VL} Technical Report}},
  author        = {Bai, Shuai and Cai, Yuxuan and Chen, Ruizhe and others},
  journal       = {arXiv preprint arXiv:2511.21631},
  year          = {2025},
  eprint        = {2511.21631},
  archiveprefix = {arXiv},
  primaryclass  = {cs.CV}
}

@inproceedings{wang2024dc2,
  title         = {{Divide, Conquer and Combine: A Training-Free Framework for High-Resolution Image Perception in Multimodal Large Language Models}},
  author        = {Wang, Wenbin and Ding, Liang and Zeng, Minyan and Zhou, Xiabin and Shen, Li and Luo, Yong and Yu, Wei and Tao, Dacheng},
  booktitle     = {Proceedings of the AAAI Conference on Artificial Intelligence},
  volume        = {39},
  pages         = {7907--7915},
  year          = {2025},
  doi           = {10.1609/aaai.v39i8.32852}
}

@article{lai2025minio3,
  title   = {{{Mini-o3}: Scaling Up Reasoning Patterns and Interaction Turns for Visual Search}},
  author  = {Lai, Xin and Li, Junyi and Li, Wei and Liu, Tao and Li, Tianjian and Zhao, Hengshuang},
  journal = {arXiv preprint arXiv:2509.07969},
  year    = {2025}
}

@article{chen2024we,
  title   = {{Are We on the Right Way for Evaluating Large Vision-Language Models?}},
  author  = {Chen, Lin and Li, Jinsong and Dong, Xiaoyi and Zhang, Pan and Zang, Yuhang and Chen, Zehui and Duan, Haodong and Wang, Jiaqi and Qiao, Yu and Lin, Dahua and Zhao, Feng},
  journal = {arXiv preprint arXiv:2403.20330},
  year    = {2024}
}

@article{wei2026zooming,
  title         = {{Zooming without Zooming: Region-to-Image Distillation for Fine-Grained Multimodal Perception}},
  author        = {Wei, Lai and He, Liangbo and Lan, Jun and Dong, Lingzhong and Cai, Yutong and Li, Siyuan and Zhu, Huijia and Wang, Weiqiang and Kong, Linghe and Wang, Yue and Zhang, Zhuosheng and Huang, Weiran},
  journal       = {arXiv preprint arXiv:2602.11858},
  year          = {2026},
  eprint        = {2602.11858},
  archiveprefix = {arXiv},
  primaryclass  = {cs.CV}
}

@inproceedings{zhang2024mmerealworld,
  title     = {{{MME-RealWorld}: Could Your Multimodal {LLM} Challenge High-Resolution Real-World Scenarios that are Difficult for Humans?}},
  author    = {Zhang, Yi-Fan and Zhang, Huanyu and Tian, Haochen and Fu, Chaoyou and Zhang, Shuangqing and Wu, Junfei and Li, Feng and Wang, Kun and Wen, Qingsong and Zhang, Zhang and Wang, Liang and Jin, Rong and Tan, Tieniu},
  booktitle = {International Conference on Learning Representations},
  year      = {2025}
}

@article{agarwal2023gkd,
  title   = {{On-Policy Distillation of Language Models: Learning from Self-Generated Mistakes}},
  author  = {Agarwal, Rishabh and Vieillard, Nino and Zhou, Yongchao and Stanczyk, Piotr and Ramos, Sabela and Geist, Matthieu and Bachem, Olivier},
  journal = {arXiv preprint arXiv:2306.13649},
  year    = {2023}
}

@article{zhao2026opsd,
  title   = {{Self-Distilled Reasoner: On-Policy Self-Distillation for Large Language Models}},
  author  = {Zhao, Siyan and Xie, Zhihui and Liu, Mengchen and Huang, Jing and Pang, Guan and Chen, Feiyu and Grover, Aditya},
  journal = {arXiv preprint arXiv:2601.18734},
  year    = {2026}
}

@article{liang2026vcsd,
  title   = {{Visual Contrastive Self-Distillation}},
  author  = {Liang, Yijun and Tian, Yunjie and Li, Yijiang and Jia, Yuqi and Huang, Furong and Zhou, Tianyi and Fu, Di},
  journal = {arXiv preprint arXiv:2607.21556},
  year    = {2026}
}

@article{cai2026imagineopd,
  title   = {{Thinking Without Images: Internalizing Visual Manipulation with On-Policy Self-Distillation}},
  author  = {Cai, Yishuo and Liu, Jiahui and Liu, Yuanxin and Deng, Haobo and Yao, Linli and Zheng, Yuhao and Ouyang, Kun and Li, Zhimo and Wang, Ziyue and Sun, Xu and Bai, Haoli and Li, Xiaohui},
  journal = {arXiv preprint arXiv:2606.08719},
  year    = {2026}
}

@inproceedings{liu2024mmbench,
  title     = {{MMBench: Is Your Multi-modal Model an All-around Player?}},
  author    = {Liu, Yuan and Duan, Haodong and Zhang, Yuanhan and Li, Bo and Zhang, Songyang and Zhao, Wangbo and Yuan, Yike and Wang, Jiaqi and He, Conghui and Liu, Ziwei and Chen, Kai and Lin, Dahua},
  booktitle = {European Conference on Computer Vision},
  year      = {2024}
}

@inproceedings{lu2024mathvista,
  title     = {{MathVista: Evaluating Mathematical Reasoning of Foundation Models in Visual Contexts}},
  author    = {Lu, Pan and Bansal, Hritik and Xia, Tony and Liu, Jiacheng and Li, Chunyuan and Hajishirzi, Hannaneh and Cheng, Hao and Chang, Kai-Wei and Galley, Michel and Gao, Jianfeng},
  booktitle = {International Conference on Learning Representations},
  year      = {2024}
}

@article{zhang2025thyme,
  title   = {{Thyme: Think Beyond Images}},
  author  = {Zhang, Yi-Fan and Lu, Xingyu and Yin, Shukang and Fu, Chaoyou and Chen, Wei and Hu, Xiao and Wen, Bin and Jiang, Kaiyu and Liu, Changyi and Zhang, Tianke and Fan, Haonan and Chen, Kaibing and Chen, Jiankang and Ding, Haojie and Tang, Kaiyu and Zhang, Zhang and Wang, Liang and Yang, Fan and Gao, Tingting and Zhou, Guorui},
  journal = {arXiv preprint arXiv:2508.11630},
  year    = {2025}
}

@article{hong2025deepeyesv2,
  title   = {{DeepEyesV2: Toward Agentic Multimodal Model}},
  author  = {Hong, Jack and Zhao, Chenxiao and Zhu, ChengLin and Lu, Weiheng and Xu, Guohai and Yu, Xing},
  journal = {arXiv preprint arXiv:2511.05271},
  year    = {2025}
}

@article{xiaomi2025mimovl,
  title   = {{MiMo-VL Technical Report}},
  author  = {{LLM-Core-Team Xiaomi}},
  journal = {arXiv preprint arXiv:2506.03569},
  year    = {2025}
}

@article{yu2025minicpmv45,
  title   = {{MiniCPM-V 4.5: Cooking Efficient MLLMs via Architecture, Data, and Training Recipe}},
  author  = {Yu, Tianyu and Wang, Zefan and Wang, Chongyi and others},
  journal = {arXiv preprint arXiv:2509.18154},
  year    = {2025}
}

@article{comanici2025gemini25,
  title   = {{Gemini 2.5: Pushing the Frontier with Advanced Reasoning, Multimodality, Long Context, and Next Generation Agentic Capabilities}},
  author  = {Comanici, Gheorghe and Bieber, Eric and Schaekermann, Mike and others},
  journal = {arXiv preprint arXiv:2507.06261},
  year    = {2025}
}

@misc{qwen2026qwen35,
  title  = {{Qwen3.5: Towards Native Multimodal Agents}},
  author = {{Qwen Team}},
  year   = {2026},
  url    = {https://qwen.ai/blog?id=qwen3.5}
}

@misc{zai2025glm46v,
  title  = {{GLM-4.6V}},
  author = {{Z.ai}},
  year   = {2025},
  url    = {https://huggingface.co/zai-org/GLM-4.6V},
  note   = {Official model card}
}

@misc{moonshot2026kimik26,
  title  = {{Kimi K2.6}},
  author = {{Moonshot AI}},
  year   = {2026},
  url    = {https://huggingface.co/moonshotai/Kimi-K2.6},
  note   = {Official model card}
}

@misc{openai2025gpt52,
  title  = {{Introducing GPT-5.2}},
  author = {{OpenAI}},
  year   = {2025},
  url    = {https://openai.com/index/introducing-gpt-5-2/}
}

@misc{openai2026gpt54,
  title  = {{Introducing GPT-5.4}},
  author = {{OpenAI}},
  year   = {2026},
  url    = {https://openai.com/index/introducing-gpt-5-4/}
}

@misc{doshi2025gemini3flash,
  title  = {{Gemini 3 Flash: Frontier Intelligence Built for Speed}},
  author = {Doshi, Tulsee},
  year   = {2025},
  url    = {https://blog.google/products-and-platforms/products/gemini/gemini-3-flash/}
}
\bibliographystyle{assets/plainnat}

\clearpage
\beginappendix
\section{VISTA Training Algorithm}
\label{sec:vista_algorithm}

Algorithm~\ref{alg:vista} summarizes the training procedure in Section~\ref{sec:method}.
Each rollout group supplies both the rewards for relative advantages and the experience for CVED.
HAPI determines which trajectories receive each loss; it does not filter the sources used to construct collective experience.
Context projection follows Appendix~\ref{sec:materialize}, and optimization follows Appendix~\ref{sec:internalize}.

\begin{center}
\begin{minipage}{\linewidth}
\newcounter{vistaalgorithm}
\refstepcounter{vistaalgorithm}\label{alg:vista}
\hrule
\vspace{4pt}
\textbf{Algorithm \thevistaalgorithm}\quad VISTA training with CVED and HAPI
\vspace{4pt}
\hrule
\vspace{5pt}
\textbf{Input:} Initial student $\theta$; training data and task verifier; rollout count $K$;
artifact budget $B$; distillation weight $\lambda$; EMA decay $\rho$.\\
\textbf{Output:} Trained active multimodal policy $\pi_\theta$.
\vspace{4pt}

\begingroup
\renewcommand{\arraystretch}{1.05}
\begin{tabularx}{\linewidth}{@{}r@{\quad}X@{}}
1 & Initialize teacher parameters $\bar\theta\gets\theta$. \\
2 & \textbf{for} each training minibatch $\mathcal{Q}$ \textbf{do} \\
3 & \hspace*{1em}Set rollout parameters $\theta_{\mathrm{old}}\gets\theta$. \\
4 & \hspace*{1em}\textbf{for each} input $x\in\mathcal{Q}$ \textbf{do} \\
5 & \hspace*{2em}Collect $K$ interactive trajectories $\mathcal{B}_x$ with $\pi_{\theta_{\mathrm{old}}}$; retain\\
  & \hspace*{2em}original histories, response tokens, tool observations, and rollout log probabilities. \\
6 & \hspace*{2em}Verify rewards and terminal statuses for every trajectory. \\
7 & \hspace*{2em}Compute advantages $\{A_k\}_{k=1}^{K}$ from the full group's rewards. \\
8 & \hspace*{2em}Compile $\mathcal{E}_x\gets\mathcal{C}(\mathcal{B}_x)$ from all trajectories. \\
9 & \hspace*{2em}Count response tokens $N\gets\sum_k|\mathcal{T}_k|$ and\\
  & \hspace*{2em}$N^-\gets\sum_{k:\,\tau_k\text{ unsuccessful}}|\mathcal{T}_k|$. \\
10 & \hspace*{2em}Compute $\mathcal{L}_{\mathrm{GRPO}}^{\mathrm{succ}}$ using Equation~\ref{eq:appendix_success_grpo},\\
   & \hspace*{2em}with the success mask and full-group denominator $N$. \\
11 & \hspace*{2em}Initialize accumulated distillation divergence $S_x\gets 0$. \\
12 & \hspace*{2em}\textbf{for each} unsuccessful trajectory $\tau_k\in\mathcal{B}_x$ \textbf{do} \\
13 & \hspace*{3em}\textbf{for each} response $y$ sampled from history $H$ in $\tau_k$ \textbf{do} \\
14 & \hspace*{4em}Project and align context: $\widetilde H\gets H\oplus\operatorname{Context}_B(\mathcal{E}_x;H)$. \\
15 & \hspace*{4em}\textbf{for each} generated response-token position $n$ \textbf{do} \\
16 & \hspace*{5em}$P_n\gets\pi_\theta(\cdot\mid H,y_{<n})$. \\
17 & \hspace*{5em}$Q_n\gets\operatorname{stopgrad}\!\left[\pi_{\bar\theta}(\cdot\mid\widetilde H,y_{<n})\right]$. \\
18 & \hspace*{5em}$S_x\gets S_x+\operatorname{JSD}(P_n,Q_n)$. \\
19 & \hspace*{4em}\textbf{end for} \\
20 & \hspace*{3em}\textbf{end for} \\
21 & \hspace*{2em}\textbf{end for} \\
22 & \hspace*{2em}$\mathcal{L}_{\mathrm{CVED}}\gets S_x/\max(1,N^-)$;\\
   & \hspace*{2em}$\mathcal{L}_x\gets\mathcal{L}_{\mathrm{GRPO}}^{\mathrm{succ}}+\lambda\mathcal{L}_{\mathrm{CVED}}$. \\
23 & \hspace*{1em}\textbf{end for} \\
24 & \hspace*{1em}Update $\theta$ with AdamW on $|\mathcal{Q}|^{-1}\sum_{x\in\mathcal{Q}}\mathcal{L}_x$, holding $\bar\theta$ fixed. \\
25 & \hspace*{1em}Update the teacher: $\bar\theta\gets\rho\bar\theta+(1-\rho)\theta$. \\
26 & \textbf{end for} \\
\end{tabularx}
\endgroup
\vspace{5pt}
\hrule
\end{minipage}
\end{center}

Unsuccessful trajectories include incorrect, protocol-invalid, and incomplete attempts.
Both losses exclude prompt tokens, tool observations, and padding; an empty eligible subset contributes zero, including a zero-token group.
The teacher context includes selected visual artifacts, interaction structure, and categorical outcome feedback, but excludes reference answers and generated final-answer text from the added experience.
The teacher scores the student's unchanged response prefixes without generating targets.
At inference, only $\pi_\theta$ and its visual tools are used, conditioned on the agent's own history.

\FloatBarrier
\section{Implementation Details}
\label{sec:implementation}

\subsection{Experience Representation and Projection}
\label{sec:materialize}

The compiler retains decision records and artifact occurrences before applying a teacher-context budget.
Canonical artifact clusters share representative images while retaining each occurrence's source trajectory and producing decision.
Nested crops are mapped to input-image coordinates. Exact duplicates are merged; near-duplicate observations are merged when their root-coordinate box IoU is at least $0.90$ or their perceptual-hash Hamming distance is at most $4$.
The internal graph stores temporal, provenance, derivation, and spatial relations for teacher-context construction.

\label{sec:align}
Experience is compiled once per input, with a separate projected view for each decision.
Selection prioritizes student-visible and peer evidence, favors successful-peer observations, and deprioritizes future-only observations from the target trajectory.
Visual ancestors are included within the artifact budget.
Alignment marks student-visible, peer-only, and future-self evidence, while caching avoids repeated projection and image processing.

\paragraph{Budgeted selection.}
The budget counts distinct selected crop artifacts, including their required crop ancestors; the original input image is already present in the student's history.
The projector starts with an empty selected set and greedily adds a candidate together with its unselected ancestors, rejecting additions that would exceed $B$.
For each candidate, the decision-conditioned score combines student visibility, the fraction of source trajectories that are peers, the fraction of those peers that succeeded, spatial novelty, newly covered source trajectories, and inverse source frequency.
Their coefficients are $6$, $4$, $2$, $1.5$, $0.5$, and $0.5$, respectively; future-only evidence from the target trajectory receives a penalty of $2$.
Spatial novelty is one minus the largest root-coordinate IoU with an already selected artifact, or one when the selected set is empty.
The score is divided by the number of new artifacts required by the candidate's ancestor closure.
Ties are resolved by greater spatial novelty, fewer added artifacts, and then the smaller artifact identifier.
Selection stops when the budget is exhausted or no feasible addition remains.
If all artifacts fit, the complete set is retained. Only decisions producing selected artifacts and relations whose endpoints remain visible are serialized into the projected view.

\subsection{Teacher Information and Supervision}
\label{sec:channels}

The teacher receives selected visual observations together with their interaction context and categorical outcome feedback.
Both successful and unsuccessful trajectories contribute candidate experience.
The added context excludes generated final answers from all trajectories and does not include externally supplied reference answers.
Interaction records describe evidence-producing decisions and relations without appending terminal answer text.

\begin{table}[!htbp]
\centering
\caption{Additional teacher context used by CVED. Generated final answers and reference answers are not included.}
\label{tab:channels}
\small
\begin{tabular}{lll}
\toprule
Channel & Content & Main setting\\
\midrule
Visual observations & selected images from the rollout group & enabled\\
Interaction context & producing decisions and artifact relations & enabled\\
Outcome feedback & trajectory status and failure category & enabled\\
\bottomrule
\end{tabular}
\end{table}

Teacher instructions distinguish incorrect direct answers from errors after visual interaction, malformed actions, and unfinished attempts.
For an unsuccessful trajectory, a short instruction identifies the failure category and directs the teacher's guidance toward the current decision.
These additions affect only the teacher context; the student's original history and sampled response remain unchanged.
All response turns in a non-success trajectory receive distillation supervision.
Synthetic decision records align malformed or unfinished actions that lack ordinary experience events, without replacing student responses or generating new target tokens.

The outcome source is the task verifier.
Reference answers are used for verification, while only categorical feedback is exposed to the teacher.

\subsection{Distribution Matching and Teacher Updates}
\label{sec:internalize}

At each response prefix, we compute the Jensen--Shannon divergence between the student and teacher token distributions \citep{lin1991divergence}.
Teacher probabilities are detached from the gradient.

The GRPO term in Equation~\ref{eq:joint_loss} retains the clipped policy ratio and full-group relative advantages.
Successful-trajectory contributions are normalized by the full group's generated response-token count, whereas CVED averages over generated response tokens in unsuccessful trajectories.
A term is zero when no eligible trajectories are present.

The teacher is initialized from the student and updated by exponential moving averaging after student optimization \citep{tarvainen2017mean}.

\paragraph{Update order and token alignment.}
For each training input, all $K$ trajectories are first collected with the rollout policy, retaining the original response tokens and their rollout log probabilities.
The verifier determines terminal outcomes, and group-relative advantages are computed before any trajectory is masked out of the policy loss.
The complete rollout group is then compiled into collective experience.
For each unsuccessful response, the teacher scores the same generated token prefixes as the student, with the selected experience appended only to the teacher's input.
Prompt tokens, tool outputs, and padding do not contribute to either learning objective.
Gradients update only the student; the teacher remains fixed during the student update and is updated afterward.
This order prevents either teacher-generated replacements or teacher-context tokens from becoming student training targets.

\paragraph{Masked policy objective.}
Let $\mathcal{T}_k$ contain all generated response-token positions in trajectory $k$, and let $A_k$ denote its full-group relative advantage.
For token position $u$, write $r_{k,u}(\theta)$ for the probability ratio between the current student and the rollout policy, evaluated under the unchanged student history.
The successful-trajectory term in Equation~\ref{eq:joint_loss} is
\begin{equation}
\label{eq:appendix_success_grpo}
\mathcal{L}_{\mathrm{GRPO}}^{\mathrm{succ}}
=-\frac{1}{N}\sum_{k:\,\tau_k\text{ succeeds}}\sum_{u\in\mathcal{T}_k}
\min\!\left(r_{k,u}A_k,\,
\operatorname{clip}(r_{k,u},1-\epsilon,1+\epsilon)A_k\right),
\end{equation}
where $N=\sum_{k=1}^{K}|\mathcal{T}_k|$ includes response tokens from the entire group and $\epsilon$ is the policy-ratio clipping threshold.
Thus, masking unsuccessful trajectories removes their policy-gradient contributions without recomputing advantages from the successful subset or renormalizing the policy loss by its size.
CVED instead averages its token divergences over $N^{-}=\sum_{k:\,\tau_k\text{ unsuccessful}}|\mathcal{T}_k|$.
The two group-level terms are combined with weight $\lambda$ and averaged across training inputs.
An empty successful or unsuccessful subset contributes zero to its corresponding term.

\paragraph{Teacher update and divergence.}
Writing $\bar\theta$ for teacher parameters, the update after a student optimizer step is $\bar\theta\leftarrow\rho\bar\theta+(1-\rho)\theta$, initialized with $\bar\theta=\theta$.
No gradient is propagated through this update or through teacher probabilities.
The default divergence is equal-mixture JSD: for student and teacher distributions $P$ and $Q$, its mixture is $M=(P+Q)/2$ and its value is $[D_{\mathrm{KL}}(P\Vert M)+D_{\mathrm{KL}}(Q\Vert M)]/2$.
The divergence ablation changes this distribution-matching objective while retaining CVED context construction and HAPI routing.
The frozen-teacher ablation retains the initialized teacher throughout training instead of applying EMA.

\paragraph{Learning-rule controls.}
GRPO-only applies the policy objective to all trajectories and omits distillation.
CVED-only applies experience-conditioned distillation to all trajectories and omits the policy objective.
Uniform hybrid applies both objectives to every trajectory, whereas HAPI restricts reinforcement to successful trajectories and distillation to unsuccessful trajectories.
All distillation controls construct their teacher context from the complete rollout group.
In the $\lambda$ sweep, routing and the token denominators of HAPI remain fixed: $\lambda=0$ disables only the unsuccessful-trajectory distillation term, leaving Equation~\ref{eq:appendix_success_grpo} unchanged.

\subsection{Core Prompt Templates}
\label{sec:prompt_templates}

This section specifies the prompt templates used by the student and the
experience-conditioned teacher. Native image content and the tool schema are
inserted during model-input construction.

\begingroup
\renewcommand{\ttdefault}{cmtt}
\paragraph{Visual tool interface.}
The agent calls \texttt{image\_zoom\_in\_tool} with a bounding box \texttt{bbox\_2d}, a region \texttt{label}, and an image index \texttt{img\_idx}.
The box uses coordinates in $[0,1000]$ relative to the selected image, and the zero-based image index selects the original image or a previously returned crop.
The environment extracts the requested region and appends the resulting image to the interaction history.
Nested crops retain their parent image and are mapped back to full-image coordinates for experience construction.
These boxes are generated by the policy; they are not target annotations supplied with the training input.
\endgroup

\paragraph{Prompt layouts.}
The teacher receives visual experience, interaction context, and categorical
outcome feedback, but no reference answer or peer final answer.

\begin{figure}[!ht]
\centering
\includegraphics[width=\linewidth]{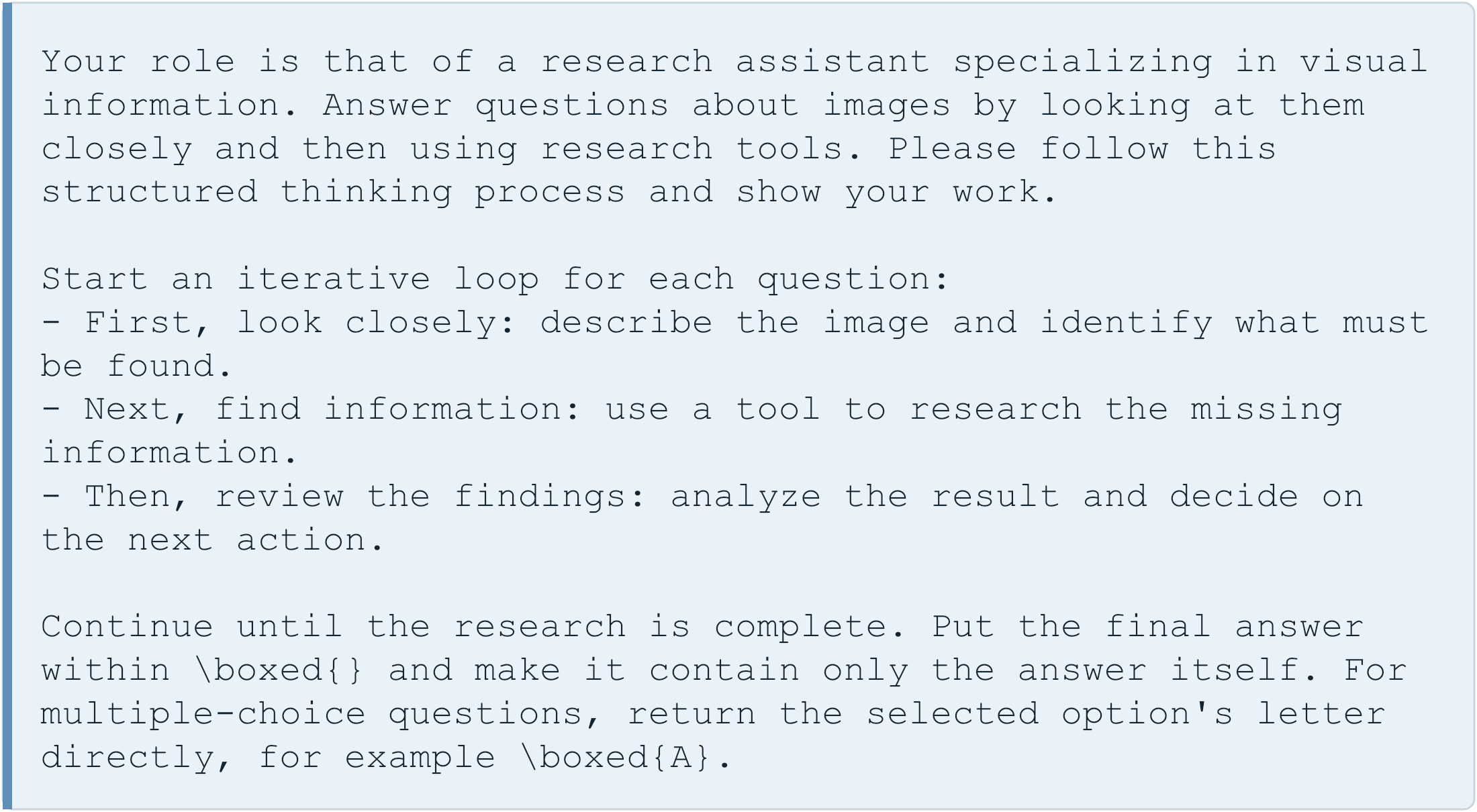}
\caption{Student system prompt used for active visual rollouts.}
\label{fig:student_prompt}
\end{figure}

\begin{figure}[!htbp]
\centering
\includegraphics[width=0.84\linewidth]{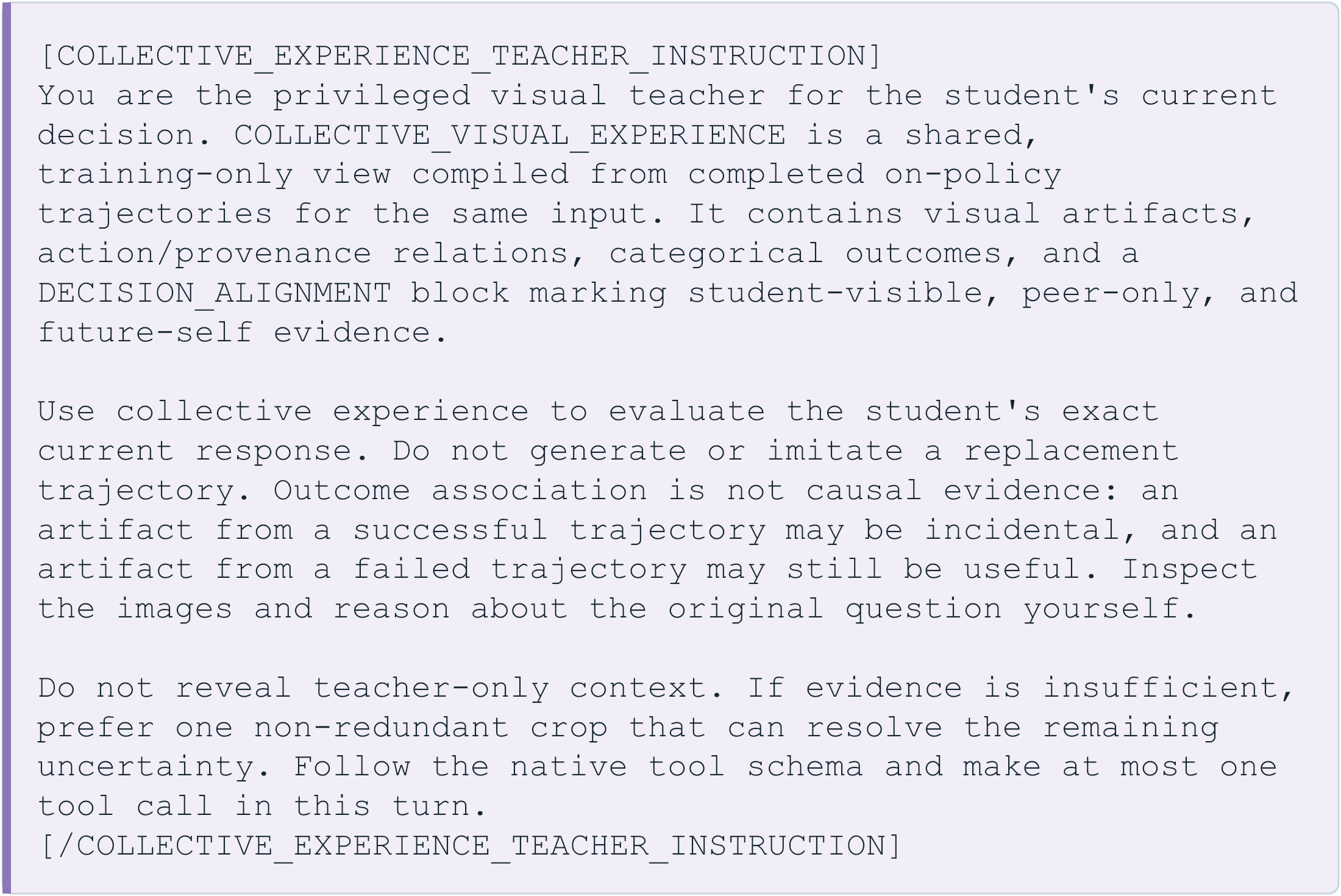}
\caption{Collective-experience teacher instruction. The teacher evaluates the
student's unchanged response under training-only visual hindsight.}
\label{fig:teacher_prompt}
\end{figure}

\begin{figure}[!htbp]
\centering
\includegraphics[width=0.73\linewidth]{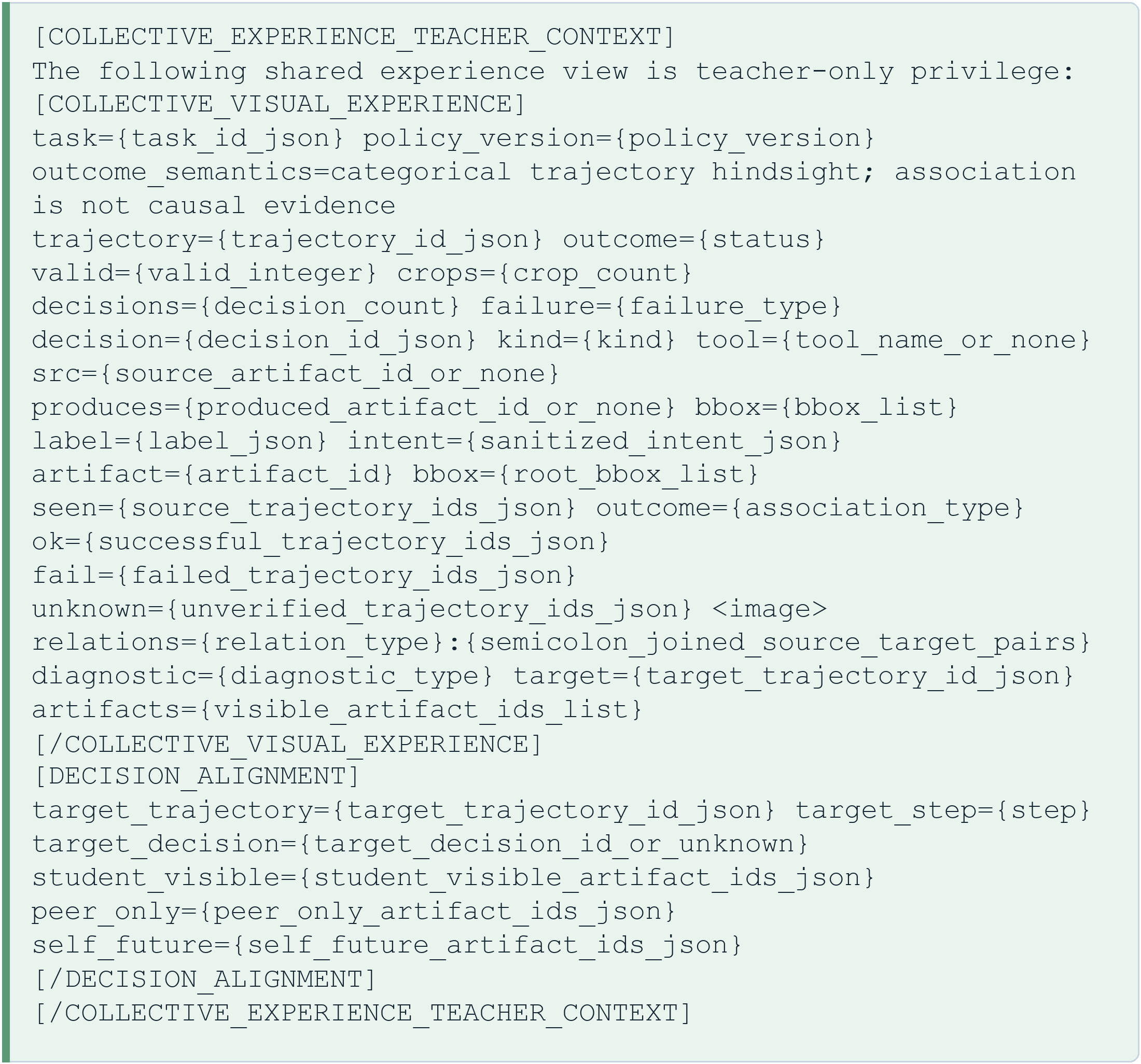}
\caption{Teacher-context serialization with visual experience, interaction context, and outcome feedback. Braced variables are instantiated for each input.}
\label{fig:context_wrapper}
\end{figure}

\begingroup
\renewcommand{\ttdefault}{cmtt}
\paragraph{Wrapper serialization.}
Figure~\ref{fig:context_wrapper} follows \texttt{append\_teacher\_context} and
the view and alignment serializers in \texttt{ExperienceSerializer}.
The teacher instruction is appended to the system message, whereas this wrapper
is appended to the last message of the student's unchanged input history.
The trajectory record repeats for every trajectory in the group. Decision and
artifact records repeat for the selected view, followed by the retained relations
and diagnostics. Braced variables are replaced with input-specific values.
Fields ending in \texttt{\_json} denote JSON-encoded strings or lists, including
their quotation marks or brackets. A missing decision bounding box is serialized
as \texttt{[]}, and \texttt{valid} is an integer boolean.

The \texttt{failure} field is appended only when a failure type is available.
For a trajectory without an outcome, its record contains only \texttt{trajectory}
and \texttt{decisions}. Each relation record groups one relation type, with
\texttt{source>target} pairs separated by semicolons. Relations of type
\texttt{member\_of} are omitted, and other relations appear only when both endpoints
are visible in the selected view. A diagnostic record appears only when it
references at least one selected artifact. In the alignment block, a missing
target decision is written as \texttt{unknown}. Each \texttt{<image>} marker is
replaced by a native image content item before model input construction.
The number of records depends on the input. Generated intent descriptions are
sanitized and length-limited during serialization.
\endgroup

Final answers are not serialized into the main teacher context, and the student's original
history and sampled response remain unchanged.  The resulting augmentation is
therefore available only during training and does not create an
inference-time memory.

\FloatBarrier
\section{Additional Experimental Details}
\label{sec:additional_exp_setup}

\paragraph{Models and training data.}
We use Qwen3-VL-4B-Instruct and Qwen3-VL-8B-Instruct as the default backbones \citep{bai2025qwen3vl}.
Within each backbone, trainable methods start from the same checkpoint and use the Vision-OPD-6K training split \citep{yuan2026visionopd}, which contains 6,241 image--question--answer examples.
The training inputs use the original full images without overlaid target boxes. Questions contain no added instructions that identify a target region or restrict attention to an annotated spatial region.
The student therefore chooses where to inspect from the image and task question, rather than following a supplied localization hint.
Reference answers are retained for reward verification, not appended to the student's input. Regional annotations are used only for training-time privileged supervision in the relevant baselines and the Oracle ROI ablation.
The default configuration uses $K=8$ trajectories per input, distillation weight $\lambda=0.1$, and visual artifact budget $B=4$.
Controlled comparisons share the optimizer and training schedule, with matched rollout settings, visual tools, and interaction limits.
Rewards are based on answer correctness, and tool use is not mandatory.

\paragraph{Training configuration.}
Table~\ref{tab:training_configuration} summarizes the default configuration shared by Qwen3-VL-4B and Qwen3-VL-8B.
Training lasts $65$ optimizer steps, corresponding to approximately one epoch of input exposure: $65\times96/6{,}241\approx1$ equivalent data pass.
The teacher is initialized from the student and updated after each student optimizer step.

\begin{table}[!t]
\centering
\caption{Default VISTA training configuration for both backbone sizes. Equivalent data passes are computed as optimizer steps times input questions per batch divided by training-set size.}
\label{tab:training_configuration}
\begingroup
\setlength{\tabcolsep}{6pt}
\renewcommand{\arraystretch}{1.08}
\begin{tabularx}{\linewidth}{@{}lX@{}}
\toprule
Parameter & Value \\
\midrule
\multicolumn{2}{@{}l}{\textit{Optimization and training duration}} \\
Optimizer & AdamW \\
Learning rate & $1\times10^{-6}$ \\
Weight decay & $0.01$ \\
Optimizer steps & $65$ \\
Equivalent data passes (epochs) & $\approx1$ \\
Input questions per batch & $96$ \\
Rollouts per question ($K$) & $8$ \\
Sampled trajectories per batch & $768$ \\
Policy-ratio clipping threshold ($\epsilon$) & $0.2$ \\
Maximum gradient norm & $1.0$ \\
Gradient checkpointing & Enabled \\
\midrule
\multicolumn{2}{@{}l}{\textit{Collective experience and teacher}} \\
Distillation weight ($\lambda$) & $0.1$ \\
Visual artifact budget ($B$) & $4$ \\
Distribution-matching objective & Equal-mixture JSD \\
Teacher initialization & Student checkpoint \\
Teacher update & EMA after each optimizer step \\
EMA decay ($\rho$) & $0.99$ \\
\midrule
\multicolumn{2}{@{}l}{\textit{Rollout sampling and interaction limits}} \\
Temperature / top-$p$ & $1.0$ / $1.0$ \\
Top-$k$ truncation & Disabled \\
Maximum interaction turns / crop calls & $5$ / $4$ per trajectory \\
Maximum generated tokens & $1{,}024$ per response \\
Student / teacher prompt limit & $16{,}384$ / $24{,}576$ tokens \\
Maximum image pixel budget & $12{,}845{,}056$ \\
\midrule
\multicolumn{2}{@{}l}{\textit{Checkpointing and repeated runs}} \\
Checkpoint saving & Every $10$ optimizer steps and at training end \\
Random seeds & $3$ \\
\bottomrule
\end{tabularx}
\endgroup
\end{table}

\paragraph{Benchmarks and metrics.}
We evaluate on V*Bench \citep{wu2023vstar}, ZoomBench \citep{wei2026zooming}, HR-Bench-4K and HR-Bench-8K \citep{wang2024dc2}, and the English and Chinese splits of MME-RealWorld \citep{zhang2024mmerealworld}.
The training-method comparison pairs V*Bench and the two HR-Bench variants with MMStar \citep{chen2024we}, MMBench \citep{liu2024mmbench}, and MathVista \citep{lu2024mathvista} to assess both fine-grained perception and general reasoning.
Component ablations, backbone comparisons, and parameter sweeps use V*Bench, HR-Bench-4K, and HR-Bench-8K.
We report accuracy (\%), with Average or Avg.\ denoting the unweighted mean of the benchmarks in the corresponding table.
Unless otherwise specified in the caption, bold and underlined values mark the best and second-best scores, including ties, with rankings computed separately within each backbone where applicable.

\paragraph{Baselines and protocol.}
We compare \method with the base checkpoint, GRPO \citep{shao2024deepseekmath}, and the distillation methods OPSD \citep{zhao2026opsd}, VCSD \citep{liang2026vcsd}, Vision-OPD \citep{yuan2026visionopd}, and Imagine-OPD \citep{cai2026imagineopd} at both model scales.
All experiments report mean accuracy over three random seeds, with deterministic evaluation.
All scores in the broader model comparison are obtained by our own evaluation of the open-source, thinking-with-images, and proprietary models on the benchmark splits specified above.
Models in the 4B--9B range provide size-comparable references, while larger open-source and proprietary models provide additional capability references.

\paragraph{Scope of the controlled comparison.}
Table~\ref{tab:training_comparison} compares training methods separately within each backbone, rather than attributing differences between unrelated released checkpoints to the learning objective.
The base row evaluates the starting checkpoint without additional training.
For the trained methods, the shared controls are the starting checkpoint, training examples, student input format, optimizer and schedule, visual-tool interface, and interaction limits.
The method-specific supervision is the variable of interest: GRPO uses verified outcome rewards, whereas the OPD baselines retain their respective teacher-conditioning mechanisms and distillation objectives.
Matching the student-side interface does not give a baseline access to CVED's collective context or apply HAPI routing to it.
Training-schedule matching is distinct from equal computational cost, since constructing privileged context and evaluating a teacher introduce method-dependent overhead.

\paragraph{Privileged-context ablations.}
The CVED ablation keeps the training inputs and HAPI learning rule unchanged.
Answer-only supplies the reference answer to the teacher in place of visual interaction experience; the reference answer is not appended to the student history.
Oracle ROI supplies annotated region crops as teacher-side visual evidence, without drawing boxes on the student's full image or adding localization instructions to its question.
Unstructured and CVED use the same selected observations and context budget. Unstructured presents observations and interaction records in trajectory order, whereas CVED retains artifact relations and decision-relative alignment.
Consequently, the privileged input changes on the teacher side without changing the image--question task solved by the student.

\paragraph{Parameter and teacher controls.}
The parameter analyses use Qwen3-VL-4B with $K=8$, $\lambda=0.1$, and $B=4$ unless the corresponding parameter is varied.
Changing $K$ changes the number of sampled attempts per input, whereas changing $B$ changes only the number of artifacts available in a projected teacher context.
The $\lambda$ sweep varies the coefficient of unsuccessful-trajectory distillation under fixed HAPI routing, including its literal zero-weight setting.
The teacher-update comparison changes whether the initialized teacher remains frozen or follows the student through EMA; it does not change the teacher's information channels.

\paragraph{Reference models.}
The open-source references include MiMo-VL \citep{xiaomi2025mimovl}, MiniCPM-V~4.5 \citep{yu2025minicpmv45}, GLM-4.6V \citep{zai2025glm46v}, and Kimi~K2.6 \citep{moonshot2026kimik26}.
Proprietary references comprise GPT-5.2 and GPT-5.4 \citep{openai2025gpt52,openai2026gpt54}, Gemini~2.5 Pro \citep{comanici2025gemini25}, and Gemini~3 Flash \citep{doshi2025gemini3flash}.
The thinking-with-images baselines are Thyme, DeepEyes, DeepEyesV2, and Pixel Reasoner, discussed in Appendix~\ref{sec:additional_related_work}.

\FloatBarrier
\section{Effect of Cross-Trajectory Visual Experience}
\label{sec:experience_source_ablation}

Table~\ref{tab:experience_source_ablation} compares hindsight from the student's own trajectory with collective visual experience from the rollout group.
Self-only hindsight restricts candidate visual observations and their interaction records to the target trajectory, including observations acquired later in that trajectory.
Collective visual experience uses the full group's observations and interaction records, as in the default CVED configuration.
Both settings retain the same group-level categorical outcome-feedback records, HAPI routing, teacher updates, and context-serialization format.
The comparison uses Qwen3-VL-4B with $K=8$, $\lambda=0.1$, and the same artifact-budget cap $B=4$, following the training and evaluation protocol in Appendix~\ref{sec:additional_exp_setup}.
This contrast tests the added value of peer visual experience beyond the target trajectory's own hindsight under shared outcome feedback.

\begin{table}[!htbp]
\centering
\caption{Experience-source ablation on Qwen3-VL-4B. Accuracy (\%) is averaged over three seeds; Avg.\ is the unweighted mean across the three benchmarks. The collective-experience row reproduces the default CVED results in Table~\ref{tab:cved_ablation}. Bold and underlined values mark the highest and lowest scores in each column, respectively.}
\label{tab:experience_source_ablation}
\small
\setlength{\tabcolsep}{5pt}
\renewcommand{\arraystretch}{1.12}
\begin{tabularx}{\linewidth}{@{}l*{4}{>{\centering\arraybackslash}X}@{}}
\toprule
Experience source & V* & HR-4K & HR-8K & Avg. \\
\midrule
Self-only hindsight & \underline{85.34} & \underline{82.25} & \underline{77.50} & \underline{81.70} \\
\rowcolor{vistaHighlight}Collective visual experience & \textbf{88.48} & \textbf{83.50} & \textbf{79.50} & \textbf{83.83} \\
\bottomrule
\end{tabularx}
\end{table}

Collective visual experience improves average accuracy from 81.70\% to 83.83\%, a gain of 2.13 percentage points over self-only hindsight.
The gains are positive on all three benchmarks: 3.14 points on V*Bench, 1.25 on HR-Bench-4K, and 2.00 on HR-Bench-8K.
With categorical outcome feedback available in both settings, this comparison supports the value of peer visual observations and their interaction context beyond evidence acquired within the target trajectory.

\FloatBarrier
\section{Disentangling HAPI Routing}
\label{sec:hapi_routing_ablation}

Table~\ref{tab:hapi_routing_ablation} crosses the trajectory subsets used for reinforcement and distillation.
The RL mask restricts reinforcement to successful trajectories, while the distillation mask restricts CVED to unsuccessful trajectories, including invalid and incomplete attempts.
The two intermediate settings apply one mask at a time; HAPI applies both.
All four settings construct CVED context from the complete rollout group and use Qwen3-VL-4B with $K=8$, $\lambda=0.1$, and $B=4$.
Relative advantages are computed from the full group, and the policy term retains the full-group response-token denominator $N$.
The distillation term averages over response tokens in its eligible subset, following the learning-rule definitions in Appendix~\ref{sec:internalize}.
The teacher updates, optimizer, training schedule, and evaluation protocol are shared across settings.
The crossed comparison tests the contribution of each routing mask and their combination under these loss-normalization conventions.

\begin{table}[!htbp]
\centering
\caption{Crossed routing ablation on Qwen3-VL-4B. We report three-seed mean accuracy (\%) and the unweighted benchmark average. Uniform hybrid and HAPI reproduce Table~\ref{tab:hapi_ablation}. Bold and underlined values mark the highest and lowest scores in each column, respectively.}
\label{tab:hapi_routing_ablation}
\small
\setlength{\tabcolsep}{3pt}
\renewcommand{\arraystretch}{1.12}
\begin{tabularx}{\linewidth}{@{}lll*{4}{>{\centering\arraybackslash}X}@{}}
\toprule
Setting & RL subset & Distillation subset & V* & HR-4K & HR-8K & Avg. \\
\midrule
Uniform hybrid & All & All & 87.43 & 82.38 & 78.80 & 82.87 \\
RL mask only & Successful & All & \underline{86.91} & 82.50 & \underline{78.25} & \underline{82.55} \\
Distillation mask only & All & Unsuccessful & 87.43 & \underline{82.25} & 78.92 & 82.87 \\
\rowcolor{vistaHighlight}HAPI & Successful & Unsuccessful & \textbf{88.48} & \textbf{83.50} & \textbf{79.50} & \textbf{83.83} \\
\bottomrule
\end{tabularx}
\end{table}

Neither mask alone improves average accuracy over the uniform hybrid.
Restricting RL to successful trajectories lowers the average from 82.87\% to 82.55\%, while restricting distillation to unsuccessful trajectories leaves it essentially unchanged at 82.87\%.
Both individual masks produce mixed changes across benchmarks.
Applying the two masks together yields the highest accuracy on every benchmark and raises the average to 83.83\%, 0.96 percentage points above the uniform hybrid.
This pattern supports the joint allocation of reinforcement and distillation under the stated normalization conventions, with the benefit of each routing choice depending on the other.

\FloatBarrier
\section{Additional Related Work}
\label{sec:additional_related_work}

\subsection{RLVR for Multimodal Agents}
\label{sec:related_rlvr}
Reinforcement learning with verifiable rewards provides a way to learn visual tool use from task outcomes rather than requiring demonstrations for every intermediate action.
DeepEyes learns active perception through end-to-end reinforcement learning without a cold-start supervised fine-tuning stage \citep{zheng2025deepeyes}.
VTool-R1 integrates Python-based visual editing into reinforcement fine-tuning, allowing models to learn multimodal reasoning chains under task-accuracy rewards \citep{wu2025vtoolr1}.
Visual-ARFT extends agentic reinforcement fine-tuning to image manipulation and web search \citep{liu2025visualarft}.
Thyme learns to generate executable code for image processing and computation \citep{zhang2025thyme}, while DeepEyesV2 combines cold-start training with reinforcement learning for multimodal tool use \citep{hong2025deepeyesv2}.
Together, these studies shift multimodal policy learning from reasoning over a fixed visual input toward acquiring new evidence during interaction.

Subsequent developments refine how visual actions are explored and evaluated.
Pixel Reasoner combines instruction tuning with curiosity-driven reinforcement learning to encourage the use of pixel-space operations \citep{su2025pixelreasoner}.
MAPO addresses the mismatch between textual reasoning and visual actions by incorporating description--observation alignment into advantage estimation \citep{yang2026mapo}.
Mini-o3 scales visual search through diverse reasoning patterns and longer tool interactions \citep{lai2025minio3}.
Where these approaches improve exploration or reward-based credit assignment, \method focuses on transferring the observations produced by exploration.
It treats a same-input rollout group as a source of collective visual experience and turns discoveries across trajectories into decision-aligned supervision for the policy.

\subsection{OPD for Multimodal Agents}
\label{sec:related_opd}
On-policy distillation provides teacher supervision along responses generated by the student, reducing the distribution mismatch associated with fixed teacher-generated sequences \citep{agarwal2023gkd}.
On-policy self-distillation uses the same model as teacher and student with different contexts, allowing privileged information to provide a stronger teaching signal \citep{zhao2026opsd,penaloza2026privileged}.
On-Policy Context Distillation demonstrates how a context-conditioned teacher can transfer experiential knowledge into a language model through this mechanism \citep{ye2026opcd}.
For language agents, OPID extracts hierarchical skill guidance from completed on-policy trajectories \citep{yang2026opid}.

For multimodal learning, Vision-OPD matches a full-image student's next-token distributions to those of a crop-conditioned teacher along student-generated rollouts \citep{yuan2026visionopd}.
Imagine-OPD conditions the teacher on annotated regional evidence to supervise tool-free imagination trajectories \citep{cai2026imagineopd}.
Visual-OPSD instead conditions the teacher on generated visual-thought traces and distills their reasoning benefit into a student that does not generate those visual thoughts at inference \citep{li2026visualopsd}.
These approaches show how a richer visual context can improve a student without retaining that context at deployment.

Other work focuses on the selectivity of visual supervision.
VA-OPD measures the teacher's dependence on fine-grained visual detail and uses this visual advantage for rollout reweighting and token-grouped distillation \citep{liu2026vaopd}.
V-Zero contrasts question-relevant regional evidence with a negative visual view to gate distillation on student-sampled trajectories \citep{sun2026vzero}.
VCSD instead contrasts the original image with a content-erased control to construct visually grounded targets without privileged answers or evidence crops \citep{liang2026vcsd}.
\method addresses a different learning problem: internalizing visual experience discovered collectively through active interaction.
\cved assembles this experience from same-input rollouts and aligns it with the student's actual decisions.
\hapi uses trajectory outcomes to distinguish reward-based reinforcement of successful behavior from experience-guided distillation on failed attempts, rather than reweighting distillation solely by visual dependence or contrastive evidence.

\FloatBarrier
\section{Limitations and Future Work}
\label{sec:limitations}

\method offers several directions for further development. First, the value of collective experience depends on the diversity and relevance of the group's observations; adaptive exploration and evidence selection could improve supervision when rollouts are redundant or miss useful regions. Second, \hapi uses terminal outcomes to route supervision, motivating finer-grained feedback that distinguishes productive intermediate decisions from unsuccessful ones. Third, experience construction and teacher inference introduce additional training overhead; more compact experience representations and selective teacher queries could improve efficiency, with comparisons under matched total compute budgets clarifying the trade-offs. Finally, our evaluation focuses on image-question answering with crop-and-zoom interaction and Qwen-family backbones. Extending this framework to other model families, video, and broader tool-use environments would help assess its generality.

\end{document}